\documentclass[letterpaper, 10 pt, conference]{ieeeconf}  %
\usepackage[letterpaper, left=48pt, right=60pt, bottom=48pt, top=55pt]{geometry}
\IEEEoverridecommandlockouts  %
\usepackage{times}
\usepackage[table]{xcolor} %
\usepackage{multicol}
\usepackage[bookmarks=true,colorlinks]{hyperref}
\usepackage{graphicx}
\usepackage{etoolbox}

\usepackage[utf8]{inputenc}
\usepackage[T1]{fontenc}
\usepackage[english]{babel}

\makeatletter
\let\NAT@parse\undefined
\makeatother
\usepackage[numbers,sort&compress]{natbib}
\usepackage{amsmath}
\usepackage{amssymb}
\usepackage{amsfonts}
\usepackage[normalem]{ulem}
\usepackage{booktabs}
\usepackage{multirow}
\usepackage{tabularx}
\usepackage{algorithm}
\usepackage{algpseudocode}
\usepackage{svg}
\usepackage{adjustbox}
\usepackage{subcaption}
\usepackage{svg}
\usepackage{overpic}

\usepackage[nameinlink,capitalise]{cleveref}
\crefname{table}{Tbl.}{Tbls.}
\Crefname{table}{Tbl.}{Tbls.}
\crefname{figure}{Fig.}{Figs.}
\Crefname{figure}{Fig.}{Figs.}
\crefname{equation}{Eq.}{Eqs.}
\Crefname{equation}{Eq.}{Eqs.}
\crefname{section}{Sec.}{Secs.}
\Crefname{section}{Sec.}{Secs.}

\newcommand{\bfx}{{\mathbf x}}
\newcommand{\bfH}{{\mathbf H}}
\newcommand{\bfI}{{\mathbf I}}
\newcommand{\R}{{\mathbb R}}
\newcommand{\SO}[1]{\mathrm{SO(#1)}}
\newcommand{\SE}[1]{\mathrm{SE(#1)}}

\definecolor{linkcolor}{rgb}{0.0,0.0,1.0}
\definecolor{purduegold}{HTML}{C28E0E} %
\hypersetup{
bookmarksopen,
bookmarksnumbered,
colorlinks=true,
allcolors=purduegold,
}

\definecolor{hicell}{HTML}{ebd99f}  %
\def\hicell{\cellcolor{hicell}}

\definecolor{grasp0}{rgb}{0.502,0.0,0.502}
\definecolor{grasp1}{rgb}{0.0,0.545,0.545}
\newcommand{\grenderwidth}{0.12\textwidth}
\newcommand{\grenderwidthlaptop}{0.13\textwidth}
\newcommand{\grenderwidthdonut}{0.12\textwidth}
\newcommand{\grenderwidthpencil}{0.12\textwidth}

\newcommand{\insertfig}{
    \includegraphics[width=\textwidth,trim={3.5cm, 6.5cm, 2cm, 0cm},clip]{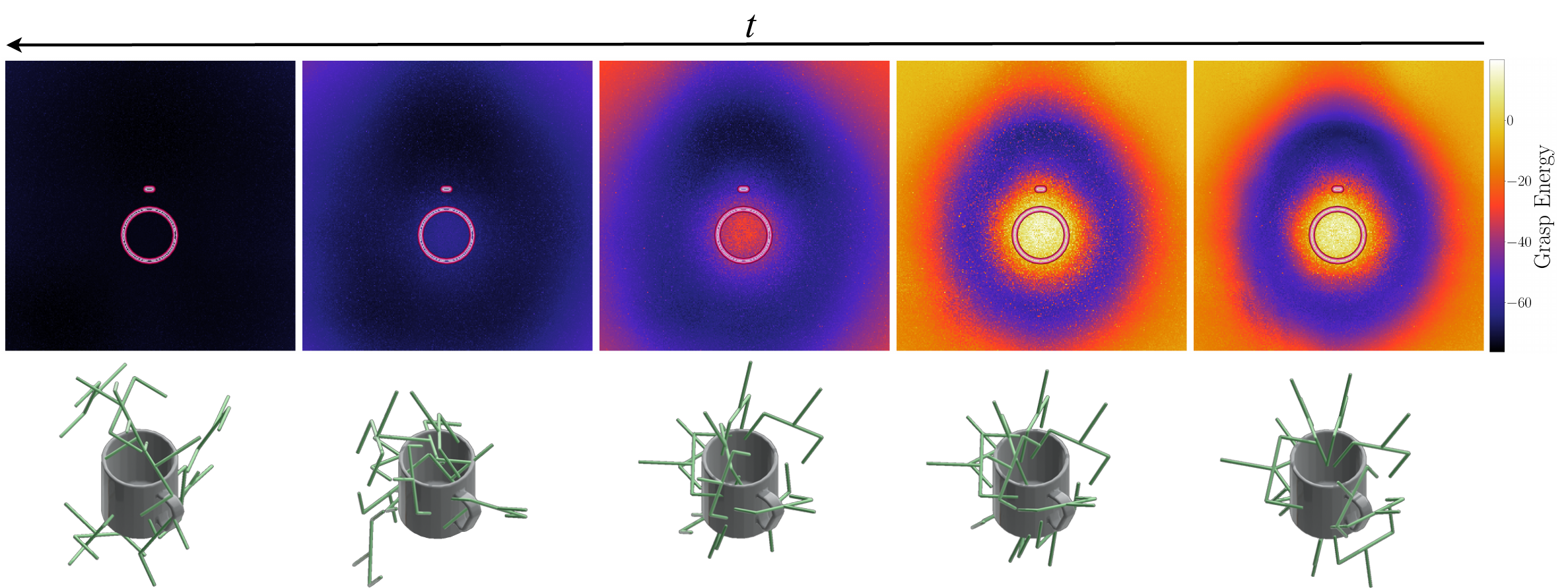}\captionof{figure}{\label{fig:teaser}
    Our energy-based formulation for Score Matching with Langevin Dynamics learns a time-conditioned grasp energy field.
    The time evolution of the learned energy field (\emph{top row}) progresses through inverse Langevin dynamics (\emph{left to right}) alongside the corresponding evolution of grasp samples (\emph{bottom row}).
    Solid lines indicate the near-zero level sets of the object.
    The energy field visualization represents the solution to the grasp orientation optimization problem evaluated at every point on a $256 \times 256$ grid.
    }
}
\makeatletter
\apptocmd{\@maketitle}{\centering\insertfig}{}{}%
\makeatother

\title{\LARGE \bf
    Variational Shape Inference for Grasp Diffusion on $\SE3$
}

\author{S. Talha Bukhari, Kaivalya Agrawal, Zachary Kingston, and Aniket Bera%
\thanks{Authors are with the Department of Computer Science, Purdue University, West Lafayette, IN 47907, USA.
\texttt{\small \{bukhars, agraw201, zkingston, aniketbera\}@purdue.edu}}
}

\begin{document}

\maketitle
\setcounter{figure}{1}  %

\begin{abstract}
Grasp synthesis is a fundamental task in robotic manipulation which usually has multiple feasible solutions.
Multimodal grasp synthesis seeks to generate diverse sets of stable grasps conditioned on object geometry, making the robust learning of geometric features crucial for success.
To address this challenge, we propose a framework for learning multimodal grasp distributions that leverages variational shape inference to enhance robustness against shape noise and measurement sparsity.
Our approach first trains a variational autoencoder for shape inference using implicit neural representations, and then uses these learned geometric features to guide a diffusion model for grasp synthesis on the $\SE3$ manifold.
Additionally, we introduce a test-time grasp optimization technique that can be integrated as a plugin to further enhance grasping performance.
Experimental results demonstrate that our \emph{shape inference for grasp synthesis} formulation outperforms state-of-the-art multimodal grasp synthesis methods on the ACRONYM dataset by 6.3\%, while demonstrating robustness to deterioration in point cloud density compared to other approaches.
Furthermore, our trained model achieves zero-shot transfer to real-world manipulation of household objects, generating 34\% more successful grasps than baselines despite measurement noise and point cloud calibration errors.
\end{abstract}

\section{Introduction}

The pursuit of autonomous robots entails sensing, planning, and interaction in the 3D world.
Due to the increase in demand for robotic operation in unstructured workspaces, meaningful object interaction has become one of the foremost goals of research in robotic manipulation~\cite{garg2020semantics}. 
Here, generating stable grasps is a key task in enabling robust manipulation in the real world.
Recent efforts for large-scale data collection have enabled data-driven approaches for robot manipulation~\cite{kim2024integrating, openxe2023}.
For learning to grasp, the primary mode of ingesting expert grasp data is via imitation learning~\cite{hsiao2006imitation, kumra2020antipodal, jiang2021synergies}.

Given an object's geometry, there are typically multiple feasible grasp solutions; hence, it is of interest to encode the \emph{multimodality} of the task and capture the entire solution set, which can then be sampled based on task-specific criteria~\cite{mousavian20196, urain2023se}.
Denoising diffusion models~\cite{song2019generative, ho2020denoising} have gained significant popularity across computer vision~\cite{dhariwal2021diffusion, rombach2022high, po2023state}, language modeling~\cite{li2022diffusion, zhang2023adding, gong2022diffuseq}, robotics~\cite{janner2022planning, carvalho2023motion, chi2023diffusion}, and beyond, due to their ability to capture highly complex multimodal distributions without mode collapse, stable training~\cite{song2021train}, and flexibility in guiding the sampling process at test-time~\cite{dhariwal2021diffusion}.
In this work, we leverage diffusion models to learn the distribution of stable grasps conditioned on object geometry.

Generating stable grasps requires understanding of object geometry to identify regions on the object where grasps are feasible.
This geometric understanding should be robust and transferable to arbitrary shapes for generalization in novel scenarios.
Thus, geometric understanding serves as a key precursor to grasp synthesis~\cite{jiang2021synergies}.
Recent advances in neural scene representations have enabled new approaches to parameterize geometry, such as implicit neural representations (INRs)~\cite{sitzmann2020implicit}.
Properties such as differentiability, smoothness, and resolution invariance with a small memory footprint make INRs attractive for robot motion planning tasks~\cite{ortiz30isdf, bukhari2025differentiable}.
The most common implicit scene representation in robotics applications is the signed distance field (SDF)~\cite{oleynikova2016signed, oleynikova2017voxblox}, which provides distance and gradients to the nearest obstacle surface.
We encode geometric features in our grasp synthesis pipeline by learning neural SDFs conditioned on object point clouds~\cite{park2019deepsdf, yang2021geometry}.

To enable a robust, generalizable, and multimodal grasp synthesis framework, we propose an approach centered around robust geometry inference.
Our pipeline uses learned geometric features for stable, multimodal grasp synthesis through a two-stage process.
We first learn a variational shape encoding guided by INRs, which maps object point clouds to a shape-latent space that enables robust shape inference using imperfect observations.
We then use the learned shape features for generating grasp poses via diffusion on the $\SE3$ manifold.
Similar to~\citet{urain2023se}, our framework can be directly incorporated into multi-objective optimization, allowing us to tune the generated grasp poses to include feasibility constraints.
We demonstrate this capability with a test-time optimization routine that uses a smooth, differentiable representation of the robot gripper to fine-tune the generated grasp poses for improved success rates.
We present results on noisy, sparse, and partial point clouds, and demonstrate the effectiveness of our method in real-world settings.
We release our code as open source\footnote{\href{https://github.com/stalhabukhari/vsigd}{\textbf{\texttt{https://github.com/stalhabukhari/vsigd}}}}.

\section{Related Work}

\subsection{Multimodal Grasp Synthesis}

Many methods have been proposed for generating a diverse set of grasps conditioned on object geometry.
\citet{mousavian20196} propose a conditional variational autoencoder (cVAE) that learns to reconstruct grasps conditioned on object point clouds.
However, VAEs are known to exhibit mode collapse and fail to fully capture the data distribution.

Diffusion models demonstrate improved multimodal distribution modeling and stable learning dynamics, avoiding mode collapse~\cite{song2021train}, and controllable sampling trade-offs and guidance mechanisms~\cite{dhariwal2021diffusion}, leading to their use in robotics~\cite{janner2022planning, carvalho2023motion}.
\citet{urain2023se}'s formulation is closest to ours, where a diffusion model is learned via denoising score matching guided by an object reconstruction cost, which represents the current state-of-the-art in multimodal grasp synthesis.
Our framework better models geometric structure in the presence of noise and calibration errors by learning to project geometric observations to a latent space of shapes and using it to infer shape features to guide grasp synthesis.

Several recent works have extended diffusion-based grasp synthesis in various directions.
\citet{chen2024behavioral} propose using stochastic interpolants to sample from an informed prior distribution instead of a Gaussian and \emph{bridge} to the target distribution.
\citet{singh2024constrained} use a part-guided diffusion approach to generate grasps tailored to areas of interest instead of the whole object.
\citet{barad2024graspldm} propose learning a latent space where diffusion on grasps is performed, and use language to condition on grasp regions.
\citet{carvalho2024grasp} propose a denoising diffusion probabilistic model for grasp synthesis, and optimize the grasps at test-time using collision spheres.
Similarly, we propose a test-time grasp pose optimization technique that boosts performance without additional training on expert grasps, demonstrating controllable generation.
\citet{lim2024equigraspflow} proposes a novel equivariant lifting layer for constructing $\SE3$-equivariant time-dependent velocity fields in flow matching.

\subsection{Neural Shape Inference}

Modern 3D shape representation methods have tackled shape inference from voxel grids~\cite{wu2015shapenets}, point clouds~\cite{qi2017pointnet}, and implicit representations~\cite{xie2022neural}.
Implicit neural representations (INRs), usually in the form of signed distance fields~\cite{park2019deepsdf, mescheder2019occupancy, sitzmann2020implicit}, have made their way into robotics tasks such as motion planning~\cite{bukhari2025differentiable}, scene understanding~\cite{ortiz30isdf}, and manipulation~\cite{quintero2024stochastic}.
Synthesizing grasps entails understanding geometric features, and hence, learning robust shape representations is key precursor to inferring stable grasp distributions conditioned on shapes.
It is desired that shape learning algorithms generalize to a wide variety of shapes and sizes~\cite{park2019deepsdf, chou2022gensdf}, and exhibit robustness to real-world noise and calibration errors.
An approach to handle noisy observations is to learn a prior over a latent space of shapes~\cite{mittal2022autosdf}.
Furthermore, using a probabilistic approach to infer shape features captures a broader set of plausible surfaces~\cite{zhou20213d, chou2023diffusion} compared to point estimates from deterministic methods.
We build on the work of~\citet{chou2023diffusion} by learned a rich latent space of shapes conditioned on object point clouds, and using the inferred shape features to drive robust grasp synthesis.

Recent work has addressed the interdependence of grasp synthesis and geometry reconstruction.
\citet{breyer2021volumetric} compute a TSDF volume of the scene, which is fed to a neural network to estimate grasp pose and quality at discrete locations on the 3D grid.
\citet{song2024implicit} propose to combine dense prediction with multimodal grasp generation, in an attempt to address the limitations of volumetric discretization.
\citet{jauhri2024learning} pose grasping as a rendering problem by jointly learning to render a locality of the object along with the grasping functions in a shared feature space.
Different from this direction of work, we consider shape learning as an implicit feature learning problem, to drive grasp synthesis.
Here, the shape inference and grasp synthesis problems can be decoupled and learned in a more scalable manner than attempting to learn jointly using expert grasp data.

\section{Background}

\begin{figure*}[t]
    \centering
    \includegraphics[width=0.93\textwidth,trim={0cm, 0.5cm, 0cm, 0.5cm},clip]{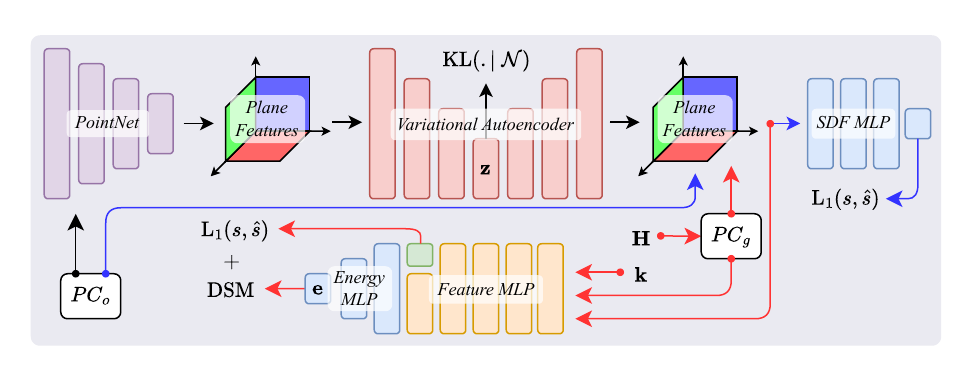}
    \caption{\label{fig:pipeline}
    Pipeline of our proposed approach. %
    We first learn variational shape encodings by training a VAE-style architecture that reconstructs planar shape features for conditioning SDF MLP queries (\emph{first row}).
    The learned shape encodings then condition the grasp diffuser by querying point cloud representations of gripper poses.
    DSM denotes the Denoising Score Matching objective and $\mathrm{L}_1$ denotes the $\mathrm{L}_1$ norm penalty.
    $PC_o$ and $PC_g$ denote object and gripper-attached point clouds, respectively.
    {\color{blue}\emph{Blue}} and {\color{red}\emph{red}} arrows indicate training procedures for the \emph{shape inference} and \emph{grasp diffusion} stages, respectively.
    }\vspace{-3mm}
\end{figure*}

\subsection{Denoising Diffusion Models}\label{ssec:dm}

Denoising diffusion models~\cite{sohl2015deep, ho2020denoising, song2021train} are a class of generative models that, given sequentially corrupted data samples (forward process), learn to reverse this corruption to generate samples from the data distribution (reverse process).
Here, we discuss the Denoising Score Matching with Langevin Dynamics (SMLD) approach to diffusion models~\cite{song2019generative}, where we learn to estimate the score of the distribution at each noise scale, and then use Langevin Dynamics to sample from progressively decreasing noise scales during generation.

Let $\rho_D(\bfx)$ denote the data distribution, and we sample $\bfx_0 \sim \rho_D(\bfx)$.
The forward process reduces sampling from the noised distribution $q_k(\bfx_k) = \int_\bfx \mathcal{N}(\bfx_k \mid \bfx, \sigma_k^2\bfI) \rho_D(\bfx) d\bfx$ to simply adding Gaussian noise to the data sample: $\bfx_k = \bfx_0 + \epsilon$ where $\epsilon \sim \mathcal{N}(\mathbf{0}, \sigma_k^2\bfI)$, $\sigma_k < \sigma_{k+1}$ for $k=1, \dots, L$, and $L$ denotes the number of noise scales.
For the reverse process, the score function $\nabla_{\bfx_k} \log q_k(\bfx_k)$ is approximated by training an estimator $s_\theta$ via the score-matching objective~\cite{songscore}:
\begin{equation}
    \frac{1}{2} \sum_{k=0}^{L} \mathbb{E} \left[ \left\| s_\theta (\bfx_k, k) - \nabla_{\bfx_k} \log \mathcal{N}(\bfx_k \mid \bfx_0, \sigma_k^2\bfI) \right\|_1 \right].
\end{equation}
The trained score model can be sampled via Annealed Langevin Dynamics~\cite{song2019generative}: $\bfx_{k-1} = \bfx_k + \frac{\alpha_k^2}{2} s_\theta(\bfx_k, k) + \alpha_k\epsilon$, where $\epsilon \sim \mathcal{N}(\mathbf{0}, \bfI)$ and $\alpha_k>0$ is a step-dependent coefficient.
SMLD effectively learns a gradient vector field pointing toward high-density regions.
An advantage of this formulation is that the score function does not depend on the generally intractable normalization constant of the underlying density function~\cite{hyvarinen2005estimation}, making it easier to evaluate.
In this work, we use SMLD to learn a generative model of grasp distributions conditioned on geometry.

\subsection{Signed Distance Fields}\label{ssec:sdf}

Signed Distance Fields (SDFs) implicitly represent geometry by providing, at every point in space, the distance to the nearest surface, with a sign denoting the interior (negative) or exterior (positive).
Formally, we let $\Omega(\bfx): \R^3 \mapsto \R$ denote the signed distance field.
The surface $\mathcal{S}$ is then represented as the zero level-set of its SDF: $\mathcal{S} = \{ \bfx \in \mathbb{R}^3 \mid \Omega(\bfx) = 0 \}$.
Its gradient satisfies the Eikonal equation: $\| \nabla_\bfx\Omega(\bfx) \| = 1$.
SDFs are differentiable almost everywhere, and where the gradient $\nabla_\bfx\Omega(\bfx)$ exists, its negative points toward the nearest surface.
In recent years, much work has focused on implicitly representing 3D geometry via neural SDFs~\cite{park2019deepsdf,ortiz30isdf}, where the SDF is parameterized by a neural network.
This allows querying at arbitrary resolutions with a small memory footprint and provides gradients via automatic differentiation.
In this work, we use SDFs to learn shape features, which we transfer to stable grasp synthesis.
We also use a learned SDF of the robot gripper for test-time pose optimization of the gripper given object point cloud.

\section{Methodology}

Our approach consists of three key components: (1) variational shape inference that learns geometric features from point clouds, (2) grasp synthesis via denoising score matching conditioned on these features, and (3) test-time pose optimization for enhanced grasp quality.

\subsection{Shape Inference via Variational Autoencoding}

High-quality grasp synthesis requires understanding geometric features that can transfer across object shapes and be robust to imperfect observations.
While inferring shape features from point clouds is an extensively researched task~\cite{erler2020points2surf, chou2022gensdf, chou2023diffusion}, multimodal frameworks that learn shape priors in a latent space offer superior robustness to observation noise and capture a broader space of plausible shape hypotheses~\cite{zhao2023michelangelo, zhou2024udiff}.
Following \citet{chou2023diffusion}, we learn SDF representations from object point clouds using a PointNet-based variational autoencoder (PointVAE) architecture that encodes useful geometric priors.
Mapping (noisy) shape measurements to the shape latent space reduces the impact of imperfect shape measurements and enforces the effect of key shape features guided by geometric priors encoded in the shape latent space.

To learn a continuous, complete, and sufficiently diverse latent space capable of modeling hundreds of shape categories and inferring their SDFs from unseen point clouds, we jointly train a conditional SDF network and a VAE.
We use the architecture of~\citet{chou2022gensdf}, which can model SDFs of thousands of objects in a unified framework.
The pipeline is comprised of a point cloud encoder $\Psi$ that generates planar features $\pi = \Psi(X)$ from the input point cloud $X \in \R^{N\times3}$.
These features are processed by a VAE, $\Theta$, to generate a latent vector $z$ and reconstructed planar features $\hat\pi$: $[\hat\pi, z] = \Theta(\pi)$.
The VAE compresses the latent space while a Gaussian prior regularizes it, enforcing continuity and completeness~\cite{chen2020learning}.

The reconstructed features $\hat\pi$ condition the SDF network $\Phi$ for predicting object-specific SDF values: $\hat{s} = \Phi(x \mid \hat\pi)$, where $x \in \R^3$ is the query point.
We learn geometric features by minimizing the $L_1$ loss on SDF values and the KL-divergence on the latent space distribution:
\begin{equation*}
    \mathcal{L}_\mathcal{S} = \| \hat{s} - s \|_1 + \beta \cdot \mathrm{KL}( q_\phi( z \mid \pi)~\|~p(z) ),
\end{equation*}
where $s \in [-1, 1]$ denotes the ground truth SDF value at $x$, $q_\phi(z \mid \pi)$ represents the inferred posterior regularized to match the prior $p(z)$ (a zero-mean Gaussian), and $\beta = 10^{-5}$ controls the regularization strength.

\subsection{Grasp Synthesis via Denoising Score Matching}

Recent works on multimodal grasp synthesis~\cite{urain2023se, carvalho2024grasp} use diffusion models to learn the distribution of stable grasps conditioned on object geometry.
Following~\citet{urain2023se}, we use Denoising Score Matching (DSM) to learn the distribution of stable grasps conditioned on object geometry, leveraging our learned shape features.

Given a grasp pose $\bfH \in \SE3$, we generate perturbed grasp poses by sampling perturbations from the Gaussian distribution on Lie groups:
\begin{equation*}
    q(\bfH \mid \bfH_\mu, \Sigma) \propto \exp \left( -0.5 \left\| \rm{Logmap}(\bfH_\mu^{-1} \bfH) \right\|_{\Sigma^{-1}}^2 \right),
\end{equation*}
where $\bfH_\mu \in \SE3$ and $\Sigma \in \R^{6\times6}$ represent the mean and covariance, respectively.
We obtain the perturbed grasp $\hat\bfH \sim q_k(\hat\bfH) = q(\hat\bfH \mid \bfH, \sigma_k \bfI)$ via:
$ \hat\bfH = \bfH \cdot \rm{Expmap}(\bf{\epsilon}) $ where $ \bf{\epsilon} \sim \mathcal{N}(\bf{0}, \sigma_k^2\bfI) $ and $0 < k \leq L$ is the noise scale.

The model learns to estimate the score $\nabla_{\hat\bfH} \log q_k(\hat\bfH)$ of the perturbed data distribution via the Score Matching objective~\cite{hyvarinen2005estimation, song2019generative}:
\begin{equation*}
    \mathcal{L}_\mathrm{DSM} = \frac{1}{L} \sum_{k=0}^{L} \mathbb{E}_{\bfH, \hat\bfH} \left[ \left\| \mathbf{s}_\theta(\hat\bfH, k) - \nabla_{\hat\bfH} \log q_k(\hat\bfH) \right\|_1 \right],
\end{equation*}
where $\mathbf{s}_\theta$ denotes the score estimator parameterized by $\theta$.

To enable composition with additional objective functions (e.g., our test-time pose optimization described in~\cref{sec:ttpo}), we formulate an energy-based model $E_\theta$ by modeling the score function as $\mathbf{s}_\theta(\hat\bfH, k) = -\nabla_{\hat\bfH} E_\theta(\hat\bfH, k)$.
Grasp synthesis proceeds by sampling $\bfH_L \sim q_L$ and running inverse Langevin dynamics in $\SE3$:
\begin{equation}
    \label{eq:ild}
    \bfH_{k-1} = \mathrm{Expmap} \left( 0.5\, \alpha_k^2 \mathbf{s}_\theta(\bfH_k, k) + \alpha_k \epsilon \right) \bfH_k,
\end{equation}
where $\epsilon \in \R^6$ is sampled from the multivariate standard normal and $\alpha_k > 0$ is a step-dependent coefficient.
By running~\cref{eq:ild} for $L$ steps, we obtain samples from the data distribution, corresponding to the grasp poses.
\Cref{fig:teaser} illustrates this process, showing the time evolution of the learned energy field during inverse Langevin dynamics along with the corresponding grasp sample evolution. 
We condition the grasp diffusion process on object geometry by querying the learned shape features $\hat\pi$ at gripper-attached point clouds, as illustrated in \cref{fig:pipeline}.
We refer the reader to~\citet{solamicro} for discussion and notation on Lie groups.

\subsection{Test-time Pose Optimization}
\label{sec:ttpo}

The energy-based formulation enables incorporating additional cost functions during inference, which is particularly valuable when reliable environmental measurements are available.
We introduce differentiable objective functions that capture intuitive grasping behaviors to drive generated grasps toward more stable poses.

We learn a neural SDF $\Omega_G$ of the robot gripper using the $L_1$ penalty on predictions and the Eikonal penalty on gradients (as described in \cref{ssec:sdf}).
Querying this neural SDF at object point clouds provides collision costs with differentiable gradients.
We compose the queried SDF values $d(X \mid \bfH)$ into a smooth differentiable collision avoidance objective:
$ \mathcal{L}_{\Omega_G}(\bfH) = \sum_X \exp \left(-\beta d(X \mid \bfH) \right) - 1 $.
To encourage stable pinching configurations, we devise an objective that minimizes deviation from optimal gripper positioning.
We use object point clouds within and near the gripper's swept volume and penalize deviation of the centroid $c_\mathrm{bbox}$ of its axis-aligned bounding box from the grasp pinching center $c_\mathrm{pinch}$:
$ \mathcal{L}_c(\bfH) = \sum_{c_\mathrm{bbox}} (c_\mathrm{bbox} - c_\mathrm{pinch})^2 $.

The combined objective $\mathcal{L}_\mathrm{opt}(\bfH) = \gamma_1 \mathcal{L}_{\Omega_G}(\bfH) + \gamma_2 \mathcal{L}_c(\bfH)$ is incorporated in the final (noise-free) stage of inverse Langevin dynamics:
\begin{equation*} 
\bfH = \mathrm{Expmap} \left( 0.5\, \alpha_0^2 \left( \mathbf{s}_\theta(\bfH, 0) + \nabla_\bfH \mathcal{L}_\mathrm{opt} ( \bfH ) \right) \right) \bfH,
\end{equation*}
where $\beta, \gamma_1, \gamma_2 > 0$ balance the influence of the objective functions with the score function.
In our experiments, we use $\beta=10, \gamma_1=200, \gamma_2=200$.
We use the SIREN architecture~\cite{sitzmann2020implicit} to model $\Omega_G$ due to its infinite differentiability and proven effectiveness for smooth, differentiable objectives~\cite{yang2021geometry}.

\begin{figure}[t]
    \centering\vspace{0.5mm}
    \begin{subfigure}{0.49\columnwidth}
        \includegraphics[width=\columnwidth, trim={140, 20, 140, 20}, clip]{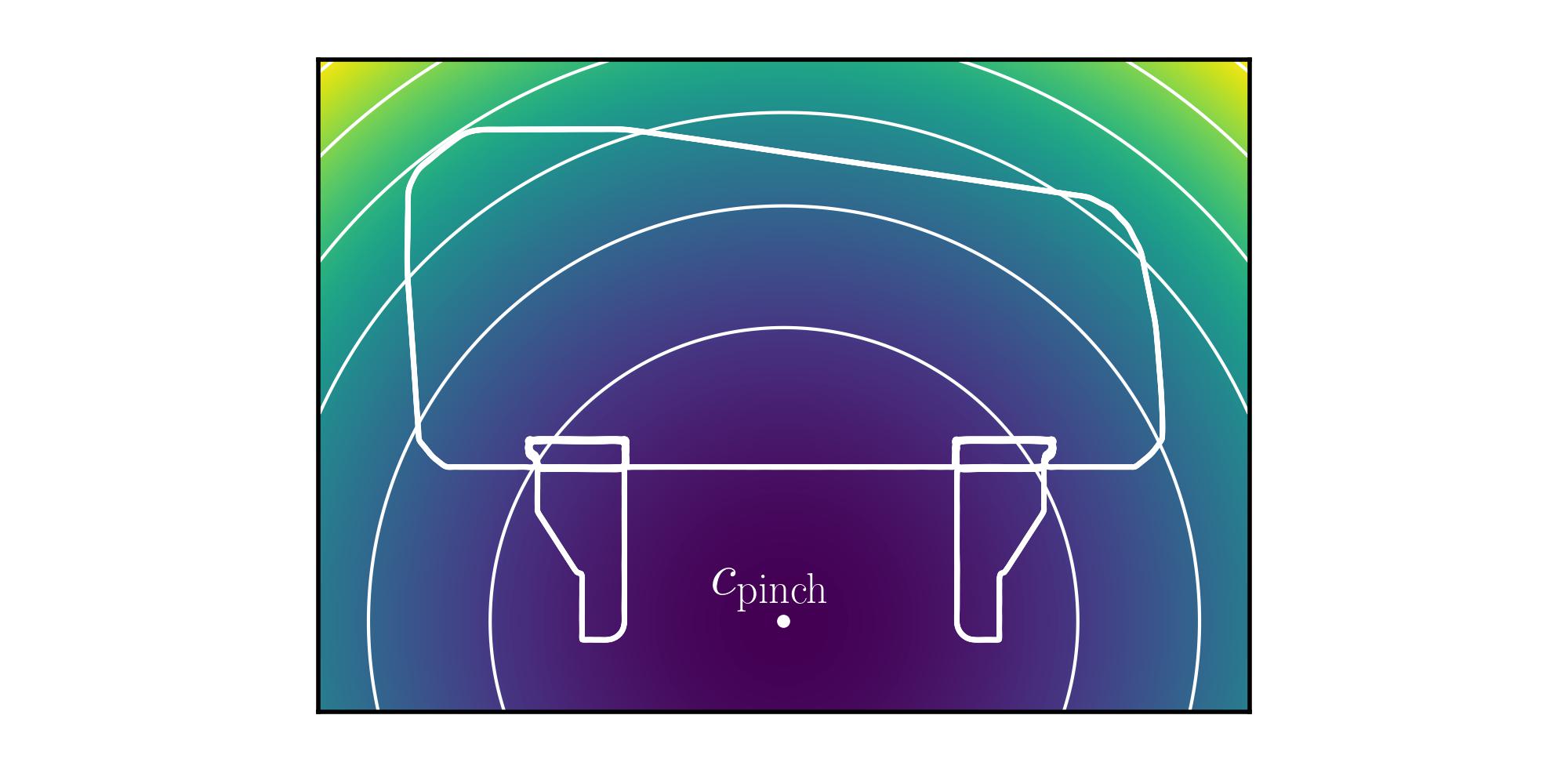}
    \end{subfigure}
    \begin{subfigure}{0.49\columnwidth}
        \includegraphics[width=\columnwidth, trim={140, 20, 140, 20}, clip]{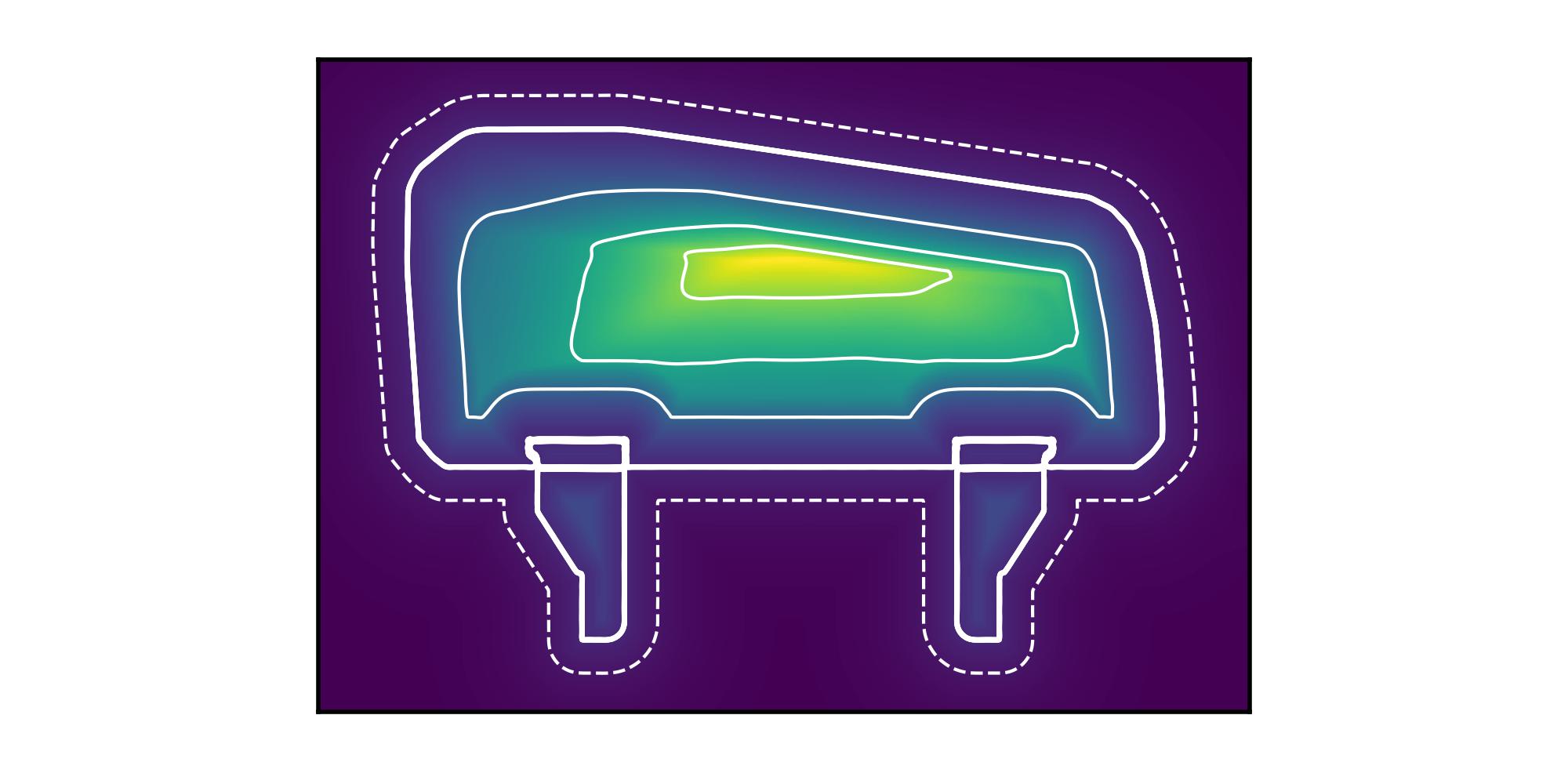}
    \end{subfigure}
    \caption{\label{fig:ttopt}
    Objective functions for test-time pose optimization: grasp pinch center alignment $\mathcal{L}_c$ (\emph{left}) and neural SDF of the gripper $\mathcal{L}_{\Omega_G}$ (\emph{right}).
    These differentiable objectives guide generated grasps toward more stable configurations during inference.
    }\vspace{-3mm}
\end{figure}

\subsection{Implementation Details}

We employ a two-stage training procedure where shape feature learning precedes grasp synthesis.
In the first stage, we train the shape autoencoder for surface reconstruction from object point clouds by coupling it with an SDF network regressor.
We use the shape autoencoder from~\citet{chou2023diffusion}, which is composed of Convolutional PointNet encoders and decoders, while the SDF network employs the SIREN architecture~\cite{sitzmann2020implicit}.

Once pretrained, we couple the shape autoencoder to the energy-based model and train for grasp synthesis through denoising score matching.
We use an MLP-style feature extractor similar to~\citet{park2019deepsdf} with the following modifications: 10 fully-connected layers interleaved with ReLU activation~\cite{nair2010rectified}, weight normalization~\cite{salimans2016weight}, and dropout~\cite{srivastava2014dropout}.
Input features are propagated to even-numbered intermediate layers via FiLM-style conditioning~\cite{perez2018film} to balance learning capacity with memory costs.
Noise-conditioning is implemented by mapping the noise level (scalar) to high-dimensional features using Gaussian Fourier Feature Projection~\cite{tancik2020fourier}.
The energy-based model comprises a two-layer ReLU-activated MLP that maps features from the feature extractor to scalar energy values.
Our coupling mechanism uses the standard data scale for the grasp diffuser while projecting data to the normalized point scale for the shape autoencoder, allowing the grasp synthesis pipeline to effectively be size-invariant.
To promote $\SO3$-equivariance, we train our method with random rotations.

\section{Experiments}

We evaluate our approach against state-of-the-art multimodal grasp synthesis frameworks: cVAE~\cite{mousavian20196}, $\SE3$-Diff~\cite{urain2023se}, Bridger~\cite{chen2024behavioral}, GLDM~\cite{barad2024graspldm}, and EGF~\cite{lim2024equigraspflow}.
All methods are trained on an NVIDIA A100 (40 GB) GPU for $\lesssim48$ hours (72 hours for partial point clouds) using ten object categories from the ACRONYM dataset~\cite{eppner2021acronym}: Book, Bottle, Bowl, Cap, CellPhone, Cup, Hammer, Mug, Scissors, and Shampoo.
We implement our method in PyTorch~\cite{pytorch2} and evaluate grasp success and stability in IsaacGym~\cite{makoviychuk2021isaac}.
Additionally, we demonstrate zero-shot sim-to-real transfer on a pick-and-place task without additional training.

\begin{table}[b]
    \centering\vspace{-2mm}
    \small  %
        \begin{tabularx}{\columnwidth}{cccc}\toprule
            \multirow[c]{2}{*}{Methods} & \multicolumn{2}{c}{Reconstruction Metrics $\downarrow$} & Failures $\downarrow$ \\ \cmidrule{2-3}
            & CD ($\times 10^{-2}$) & UHD ($\times 10^{-2}$) & ($\cdot / 7897$) \\
            \midrule
            VNN~\cite{deng2021vector} & $11.07$~\scriptsize{$\pm 06.85$} & $23.43$~\scriptsize{$\pm 18.25$} & 42 \\
            \midrule
            PointVAE~\cite{chou2023diffusion} & $08.65$~\scriptsize{$\pm 03.50$} & $17.24$~\scriptsize{$\pm 12.67$} & 4 \\
            \bottomrule
        \end{tabularx}
    \caption{\label{table:shape-infer}
    Shape inference performance comparing geometry reconstruction from point clouds.
    CD denotes \emph{Chamfer Distance} and UHD denotes \emph{Unidirectional Hausdorff Distance}.
    \emph{Failures} indicates the number of failed surface extractions by the Marching Cubes algorithm~\cite{lorensen1998marching}.
    }
\end{table}

\subsection{Shape Inference}
\label{ssec:recon}

Effective grasp synthesis requires robust geometric features that transfer across object shapes and remain stable under imperfect observations.
To validate that our shape inference architecture (PointVAE) provides such features, we evaluate its surface reconstruction capabilities against a baseline approach.
We train both our PointVAE-based encoder and a Vector Neurons~\cite{deng2021vector} (VNN) encoder coupled with an MLP using an $L_1$ loss for object reconstruction on meshes from the ACRONYM dataset.

As shown in \cref{table:shape-infer}, the PointVAE backbone achieves superior reconstruction quality with lower Chamfer Distance (CD) and Unidirectional Hausdorff Distance (UHD) compared to the VNN encoder.
The significantly lower failure rate in reconstructing the zero-level set with Marching Cubes surface extraction (4 vs. 42 failures) demonstrates that PointVAE better captures surface geometry.
Crucially, learning a latent space of object geometries enables robust shape inference from limited observations, as we demonstrate in our point cloud sparsity experiments (\cref{ssec:pc-sparse}).

\subsection{Grasp Synthesis}\label{exp:grasp}

\begin{figure}[t]
    \centering\vspace{-1mm}
    \hspace{-4mm}
    \begin{subfigure}{0.49\columnwidth}
        \includeinkscape[width=1.09\columnwidth]{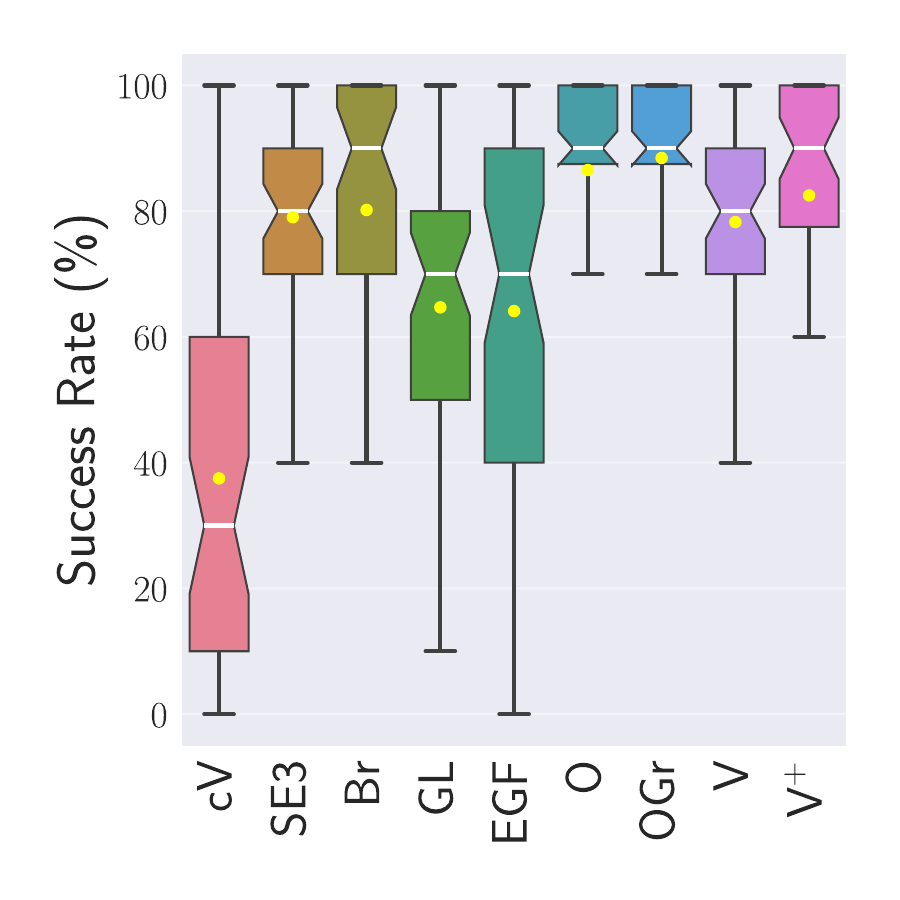}
        \label{fig:boxplots-sr}
    \end{subfigure}
    \begin{subfigure}{0.49\columnwidth}
        \includeinkscape[width=1.09\columnwidth]{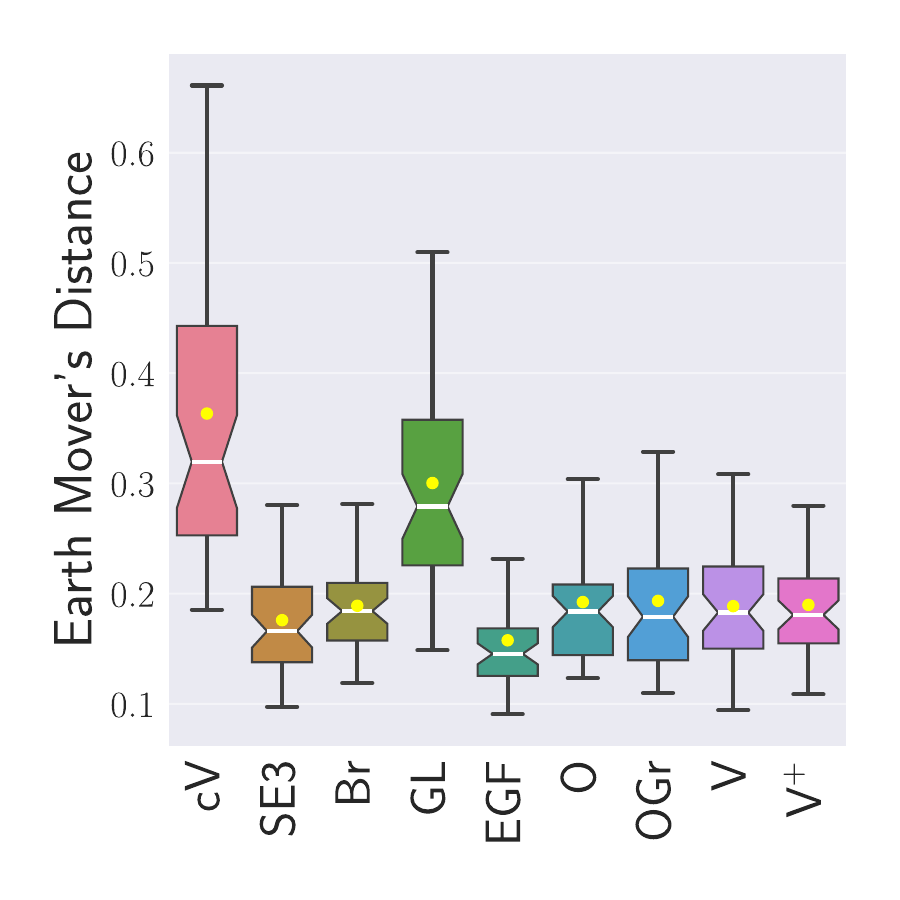}
        \label{fig:boxplots-emd}
    \end{subfigure}
    \newline\vspace{-11mm}
    \begin{subfigure}{\columnwidth}
        \includeinkscape[width=\columnwidth]{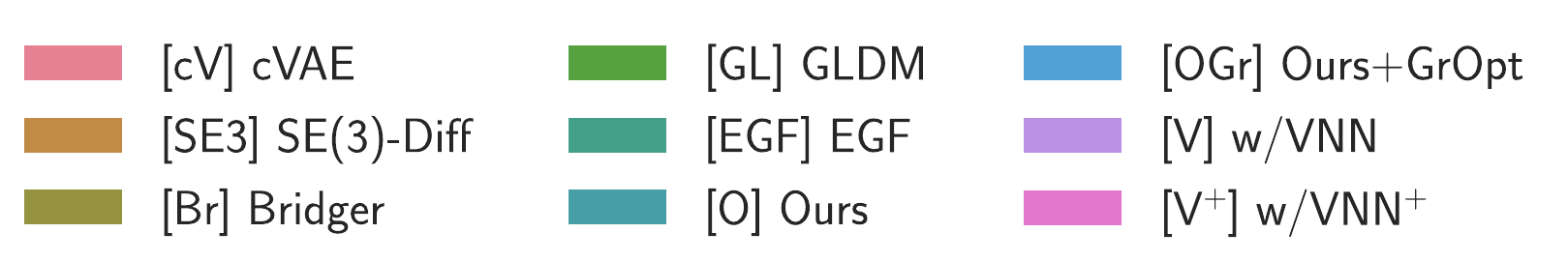}
        \label{fig:boxplots-legend}
    \end{subfigure}
    \vspace{-8mm}
    \caption{\label{fig:boxplots}
    Distribution of performance metrics.
    Our method demonstrates the most consistent grasp performance across test objects while maintaining competitive grasp diversity.
    }
    \vspace{-3mm}
\end{figure}

\begin{figure}[b]
    \centering\vspace{-3.5mm}
    \hspace{-4mm}
    \begin{subfigure}{0.49\columnwidth}
        \includeinkscape[width=1.09\columnwidth]{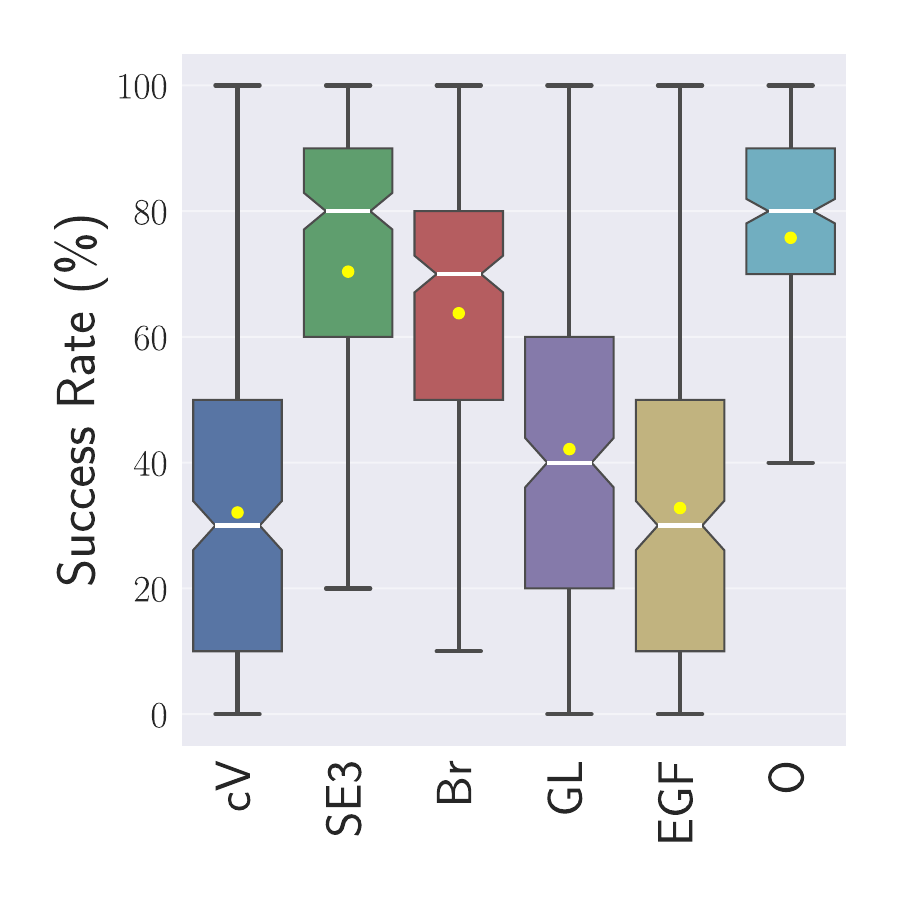}
        \label{fig:boxplots-sr}
    \end{subfigure}
    \begin{subfigure}{0.49\columnwidth}
        \includeinkscape[width=1.09\columnwidth]{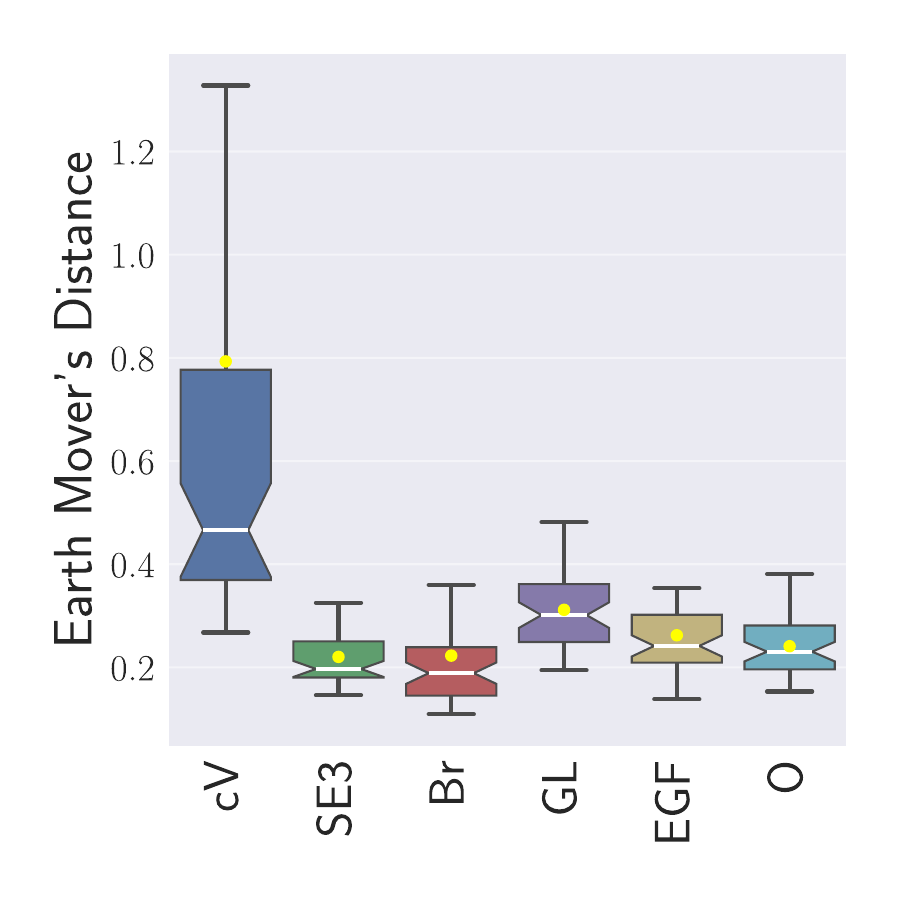}
        \label{fig:boxplots-emd}
    \end{subfigure}
    \newline\vspace{-11mm}
    \begin{subfigure}{\columnwidth}
        \includeinkscape[width=\columnwidth]{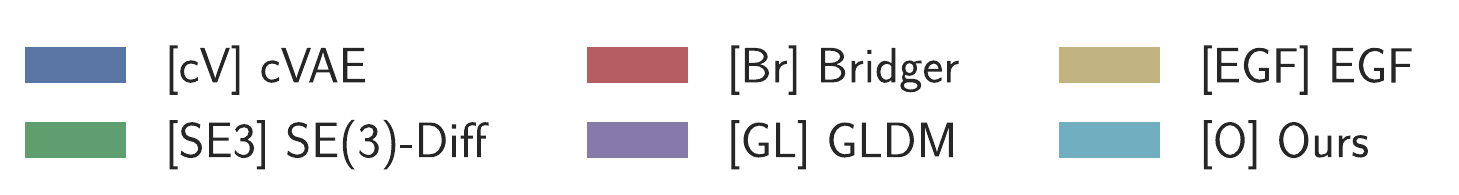}
        \label{fig:boxplots-legend}
    \end{subfigure}
    \vspace{-8mm}
    \caption{\label{fig:boxplots-partial}
    Performance evaluation on partial point clouds.
    Our method demonstrates the most consistent grasp performance while maintaining diversity across test objects despite using single-view measurements only.
    }
\end{figure}

\begin{figure*}[t]
    \centering
    \begin{subfigure}{\grenderwidth}
        \caption{cVAE~\cite{mousavian20196}}
    \end{subfigure}
    \hfill
    \begin{subfigure}{\grenderwidth}
        \caption{$\SE3$-Diff~\cite{urain2023se}}
    \end{subfigure}
    \hfill
    \begin{subfigure}{\grenderwidth}
        \caption{Bridger~\cite{chen2024behavioral}}
    \end{subfigure}
    \hfill
    \begin{subfigure}{\grenderwidth}
        \caption{GLDM~\cite{barad2024graspldm}}
    \end{subfigure}
    \hfill
    \begin{subfigure}{\grenderwidth}
        \caption{EGF~\cite{lim2024equigraspflow}}
    \end{subfigure}
    \hfill
    \begin{subfigure}{\grenderwidth}
        \caption{\textbf{Ours}}
    \end{subfigure}
    \hfill
    \begin{subfigure}{\grenderwidth}
        \caption{\textbf{Ours + GrOpt}}
    \end{subfigure}
    \newline
    \begin{subfigure}{\grenderwidthlaptop}
        \begin{overpic}[width=\textwidth,tics=10,trim=100 100 100 40,clip]{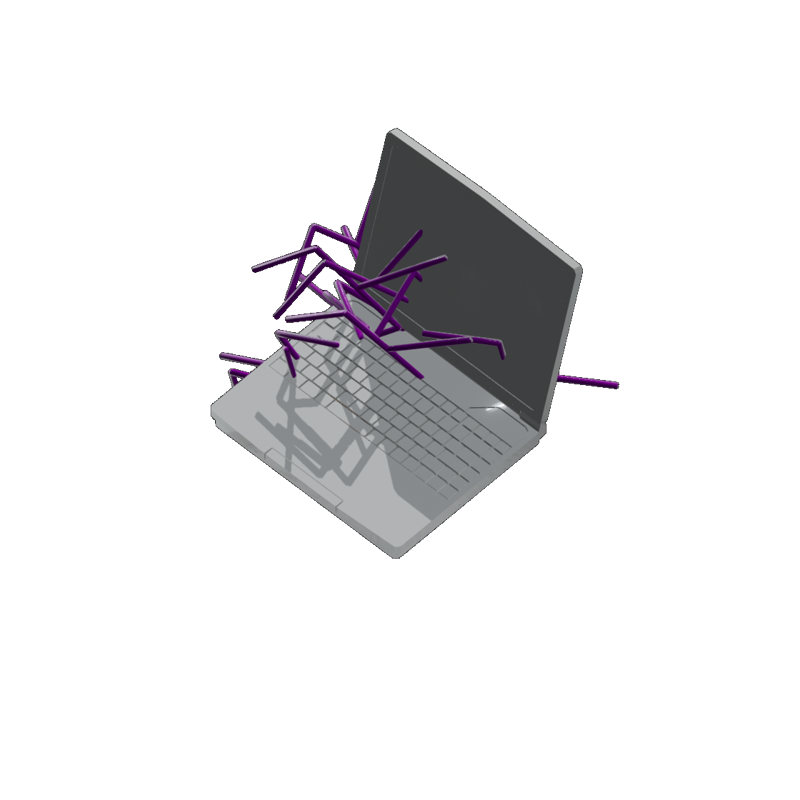}
        \end{overpic}
    \end{subfigure}
    \hfill
    \begin{subfigure}{\grenderwidthlaptop}
        \begin{overpic}[width=\textwidth,tics=10,trim=100 100 100 40,clip]{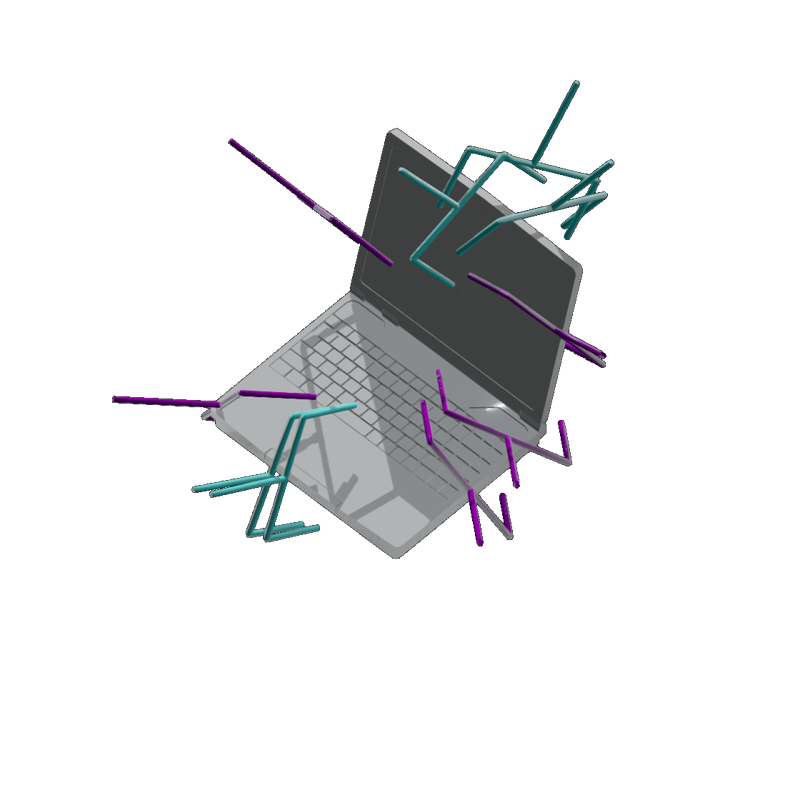}
        \end{overpic}
    \end{subfigure}
    \hfill
    \begin{subfigure}{\grenderwidthlaptop}
        \begin{overpic}[width=\textwidth,tics=10,trim=100 100 100 40,clip]{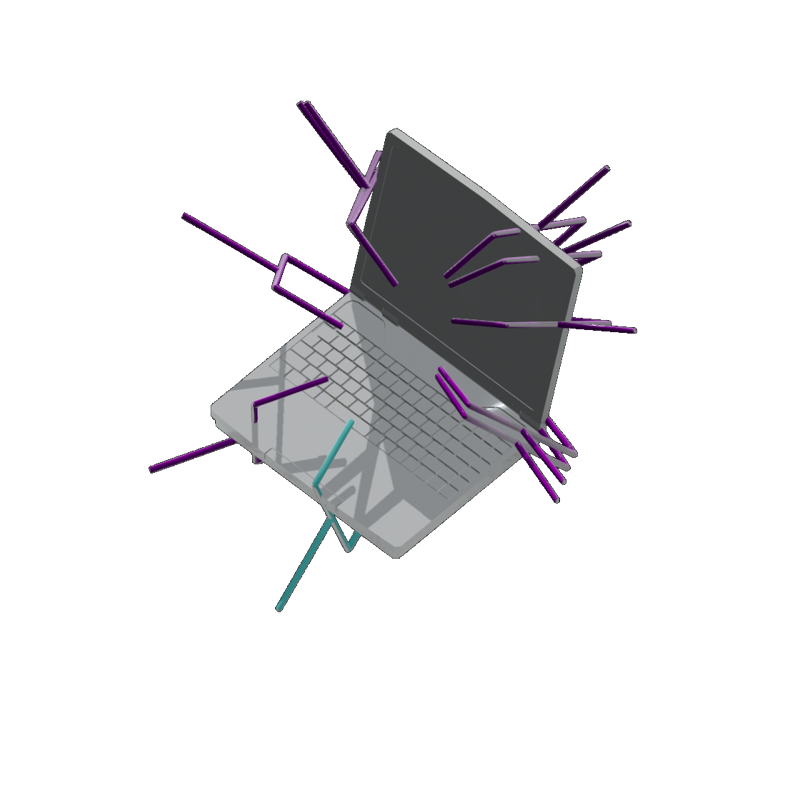}
        \end{overpic}
    \end{subfigure}
    \hfill
    \begin{subfigure}{\grenderwidthlaptop}
        \begin{overpic}[width=\textwidth,tics=10,trim=100 100 100 40,clip]{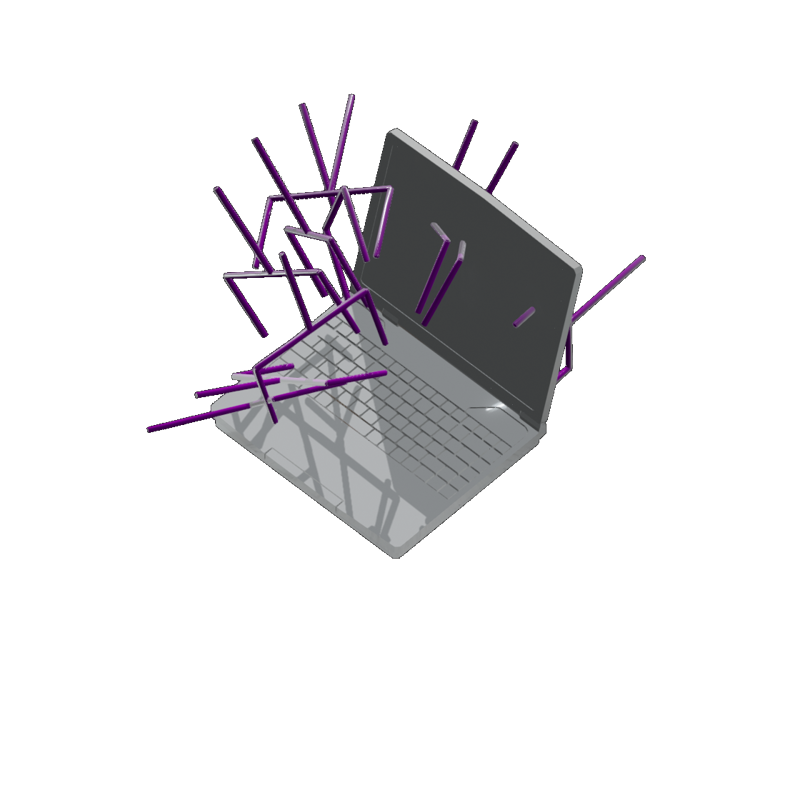}
        \end{overpic}
    \end{subfigure}
    \hfill
    \begin{subfigure}{\grenderwidthlaptop}
        \begin{overpic}[width=\textwidth,tics=10,trim=100 100 100 40,clip]{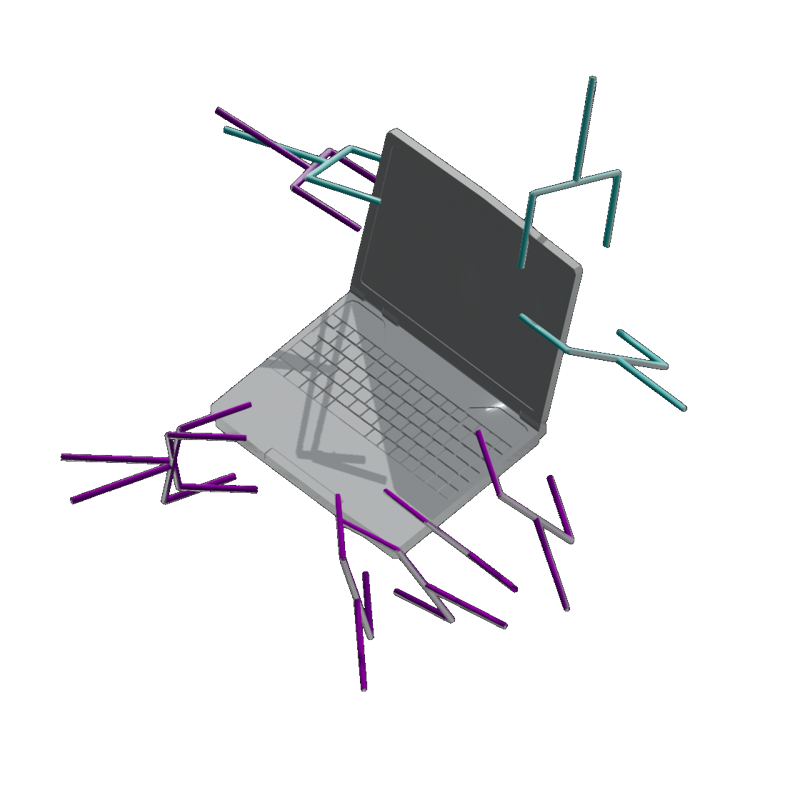}
        \end{overpic}
    \end{subfigure}
    \hfill
    \begin{subfigure}{\grenderwidthlaptop}
        \begin{overpic}[width=\textwidth,tics=10,trim=100 100 100 40,clip]{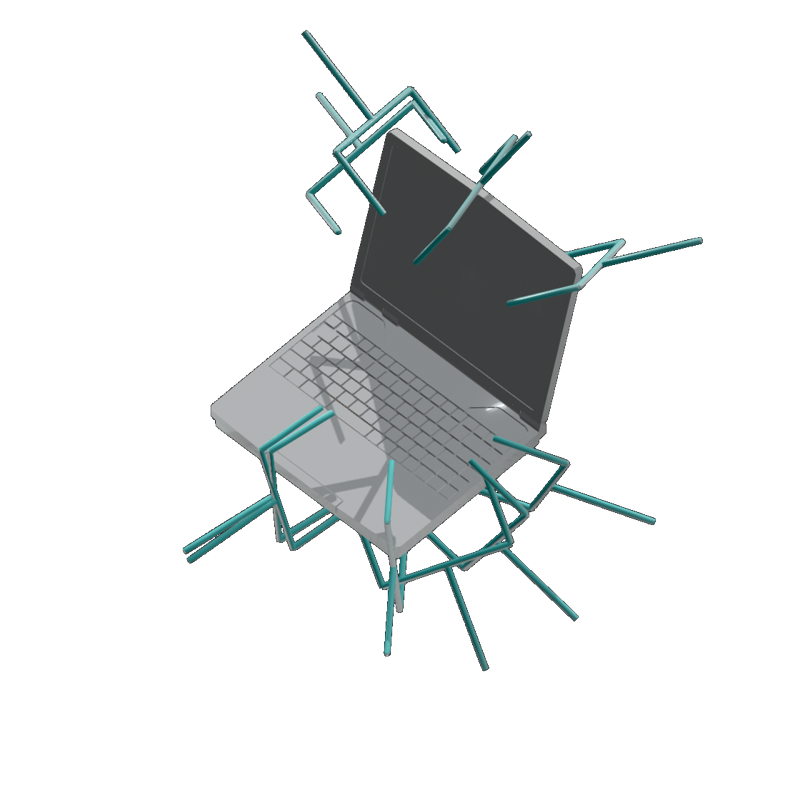}
        \end{overpic}
    \end{subfigure}
    \hfill
    \begin{subfigure}{\grenderwidthlaptop}
        \begin{overpic}[width=\textwidth,tics=10,trim=100 100 100 40,clip]{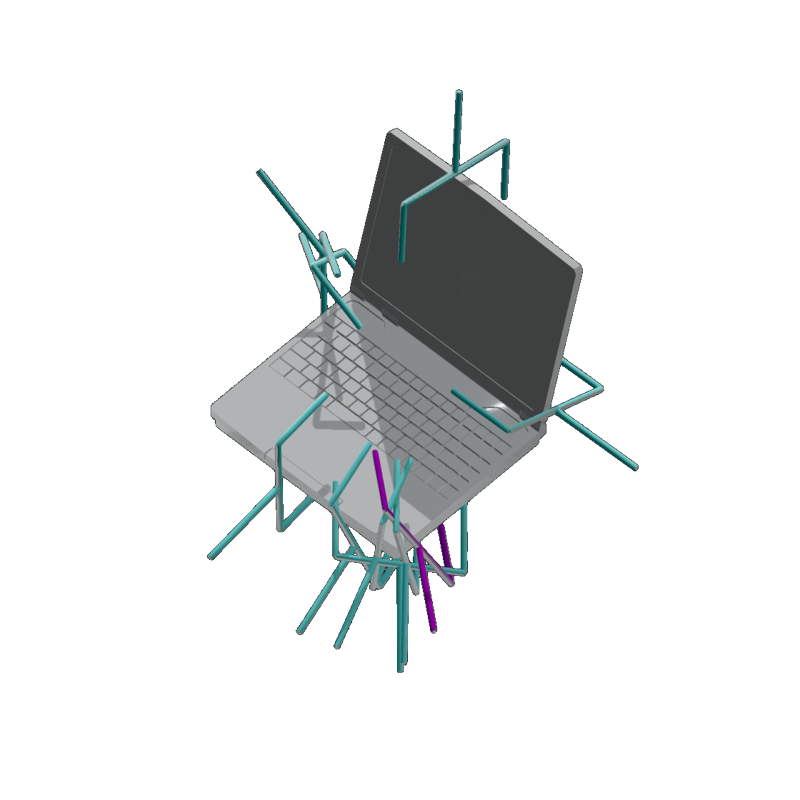}
        \end{overpic}
    \end{subfigure}
    \newline
    \begin{subfigure}{\grenderwidthdonut}
        \begin{overpic}[width=\textwidth,tics=10,trim=30 100 170 140,clip]{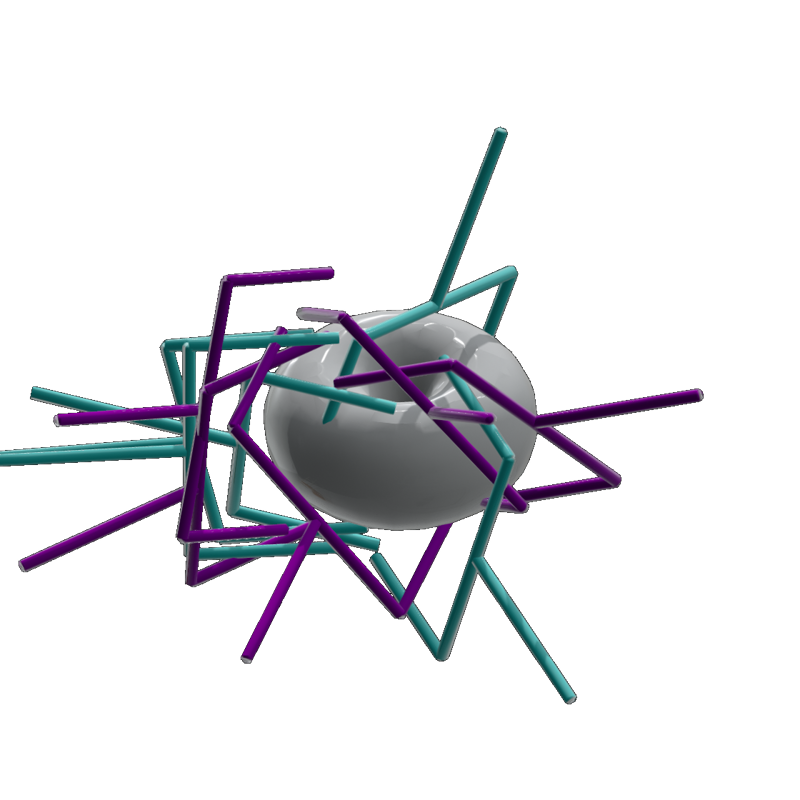}
        \end{overpic}
    \end{subfigure}
    \hfill
    \begin{subfigure}{\grenderwidthdonut}
        \begin{overpic}[width=\textwidth,tics=10,trim=100 100 100 140,clip]{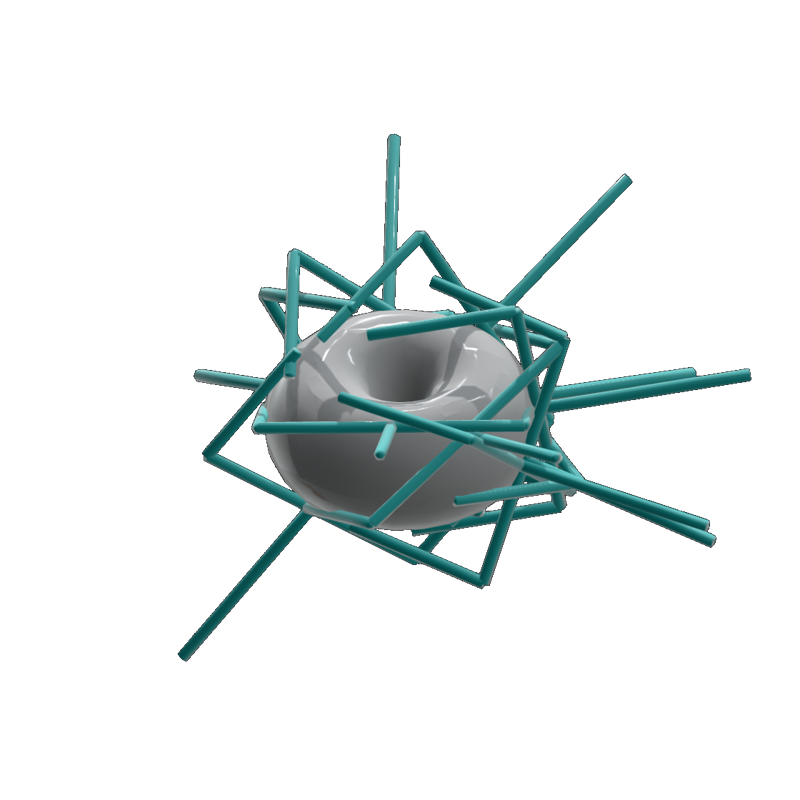}
        \end{overpic}
    \end{subfigure}
    \hfill
    \begin{subfigure}{\grenderwidthdonut}
        \begin{overpic}[width=\textwidth,tics=10,trim=100 100 100 140,clip]{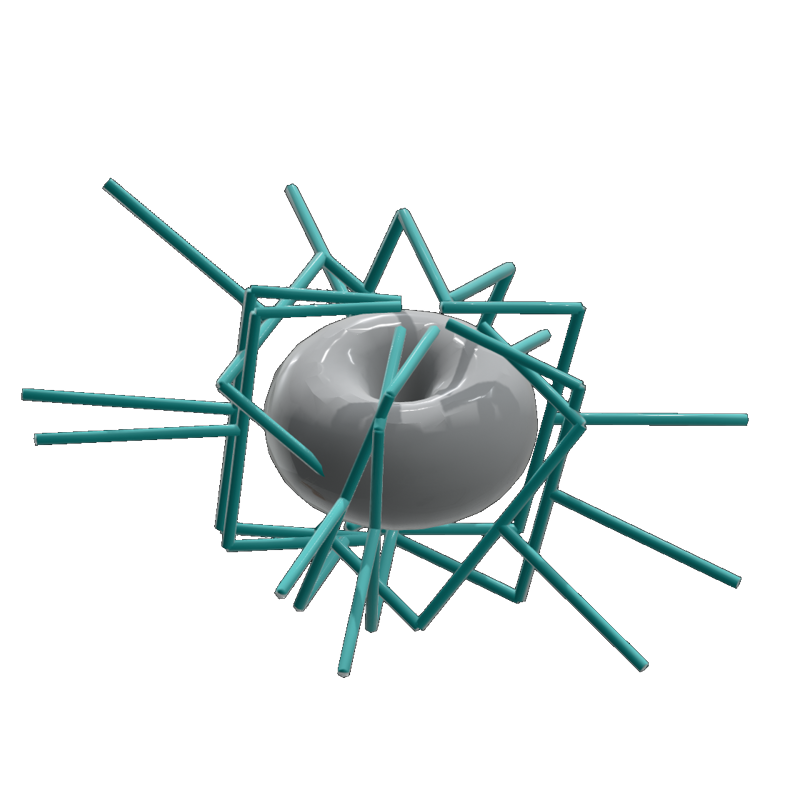}
        \end{overpic}
    \end{subfigure}
    \hfill
    \begin{subfigure}{\grenderwidthdonut}
        \begin{overpic}[width=\textwidth,tics=10,trim=100 100 100 140,clip]{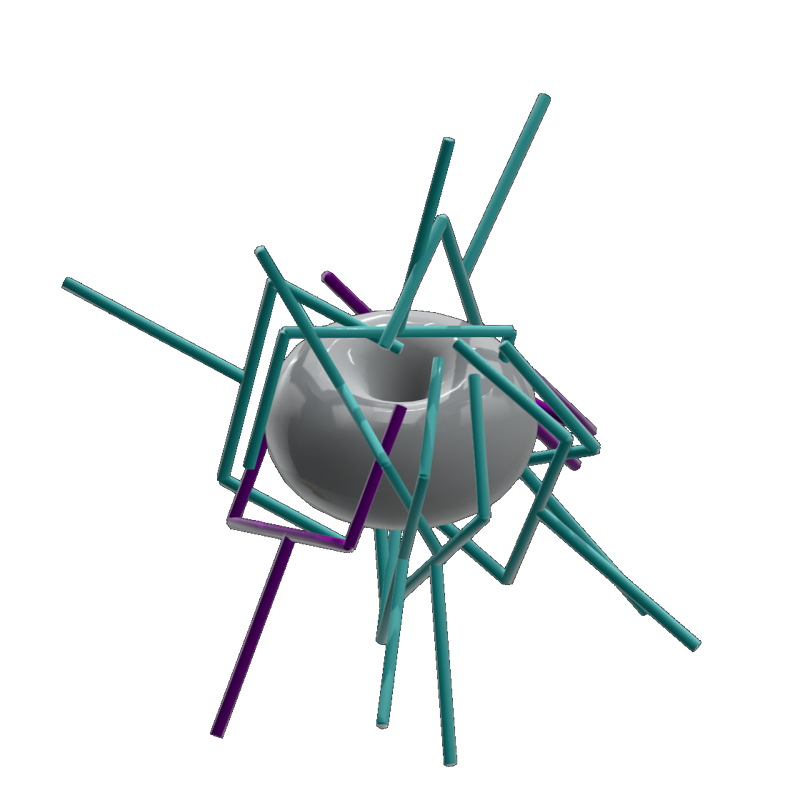}
        \end{overpic}
    \end{subfigure}
    \hfill
    \begin{subfigure}{\grenderwidthdonut}
        \begin{overpic}[width=\textwidth,tics=10,trim=100 100 100 140,clip]{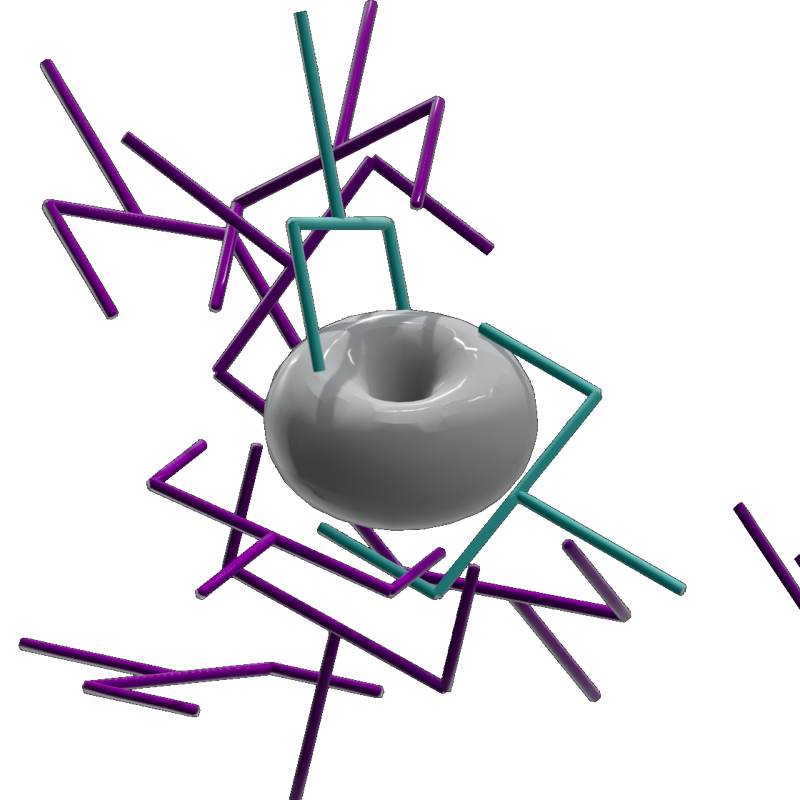}
        \end{overpic}
    \end{subfigure}
    \hfill
    \begin{subfigure}{\grenderwidthdonut}
        \begin{overpic}[width=\textwidth,tics=10,trim=100 100 100 140,clip]{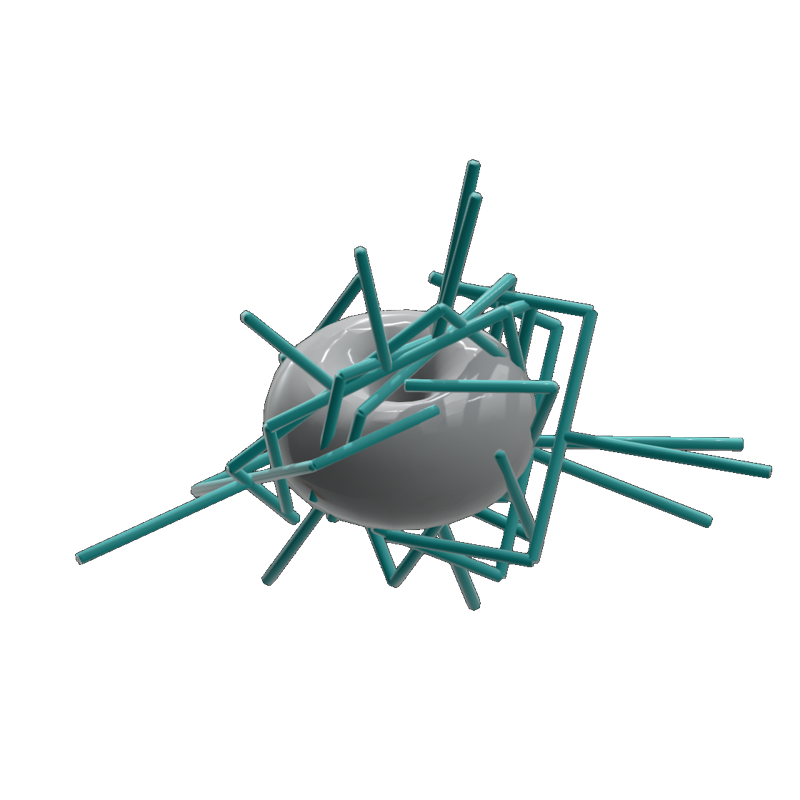}
        \end{overpic}
    \end{subfigure}
    \hfill
    \begin{subfigure}{\grenderwidthdonut}
        \begin{overpic}[width=\textwidth,tics=10,trim=100 100 100 140,clip]{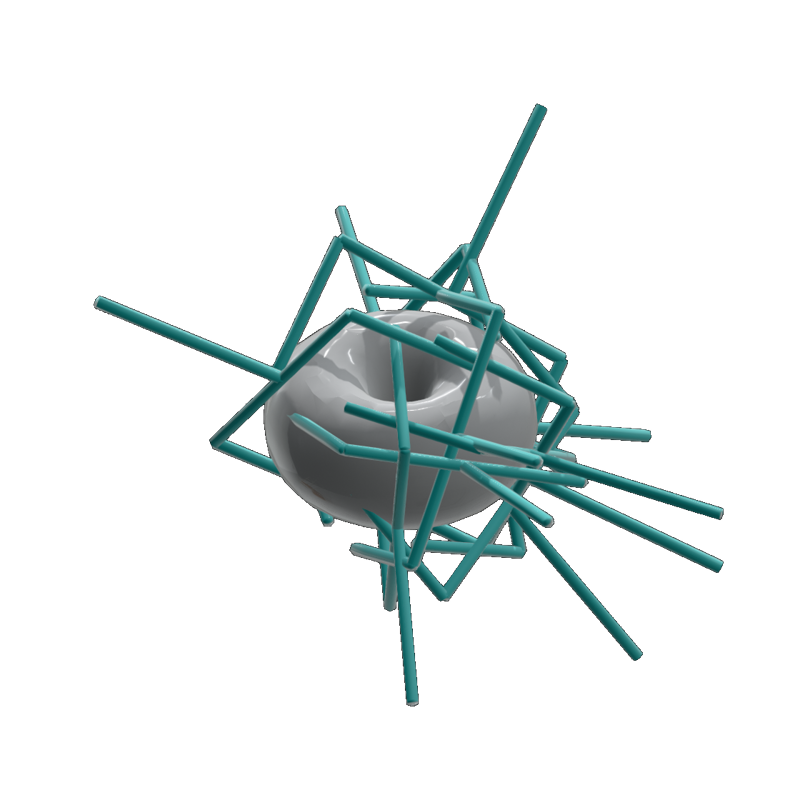}
        \end{overpic}
    \end{subfigure}
    \newline
    \begin{subfigure}{\grenderwidthpencil}
        \begin{overpic}[width=\textwidth,tics=10,trim=100 120 140 50,clip]{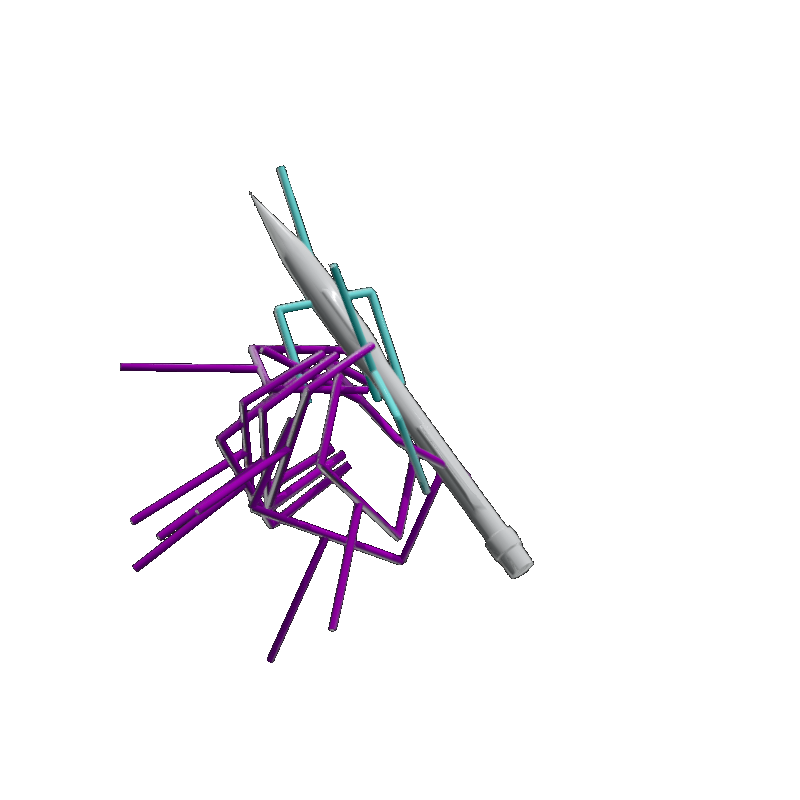}
        \end{overpic}
    \end{subfigure}
    \hfill
    \begin{subfigure}{\grenderwidthpencil}
        \begin{overpic}[width=\textwidth,tics=10,trim=120 120 120 50,clip]{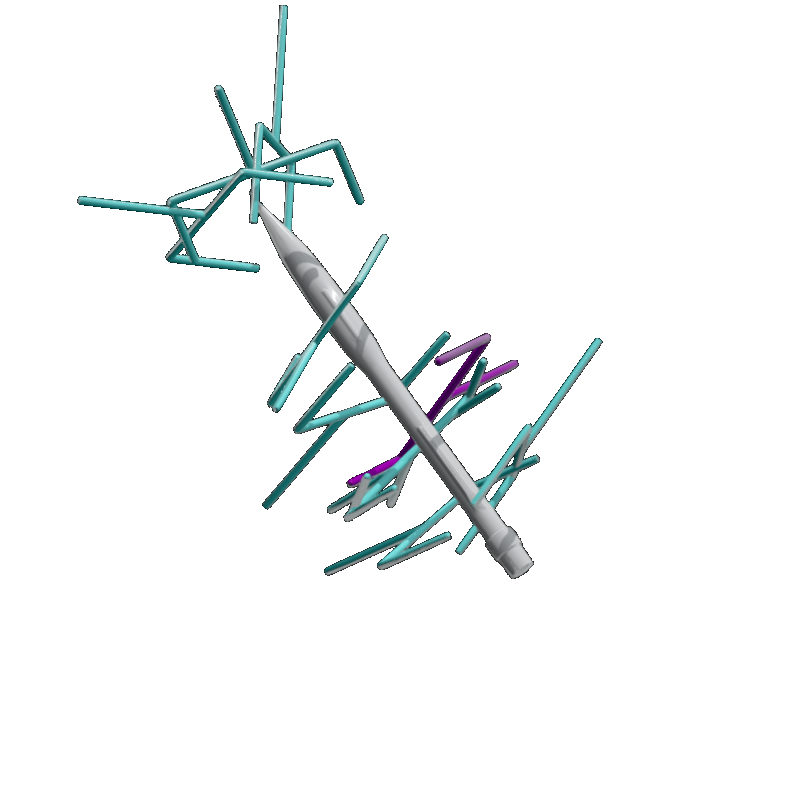}
        \end{overpic}
    \end{subfigure}
    \hfill
    \begin{subfigure}{\grenderwidthpencil}
        \begin{overpic}[width=\textwidth,tics=10,trim=120 120 120 50,clip]{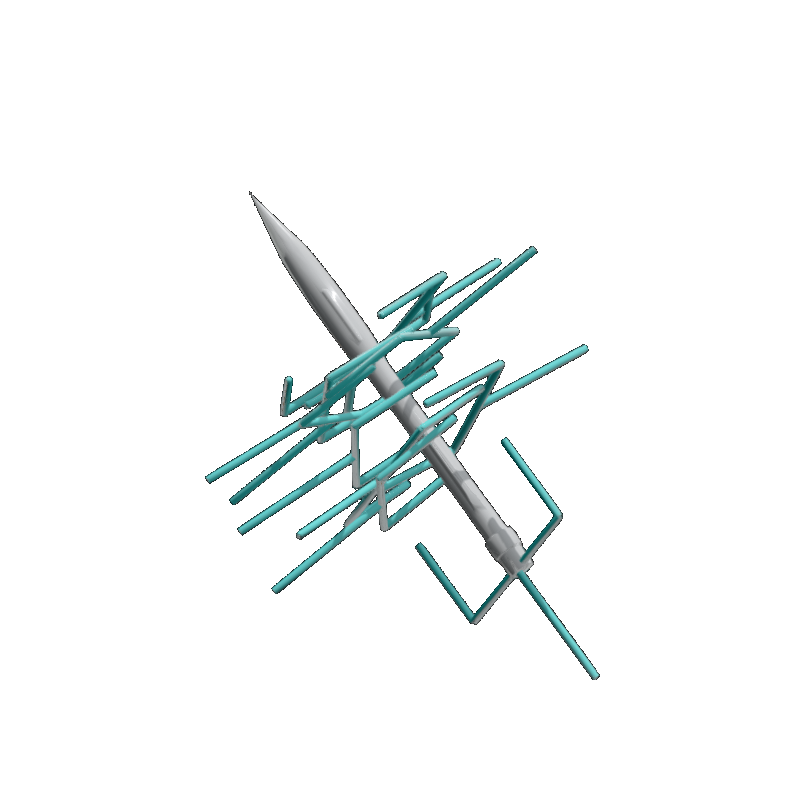}
        \end{overpic}
    \end{subfigure}
    \hfill
    \begin{subfigure}{\grenderwidthpencil}
        \begin{overpic}[width=\textwidth,tics=10,trim=120 120 120 50,clip]{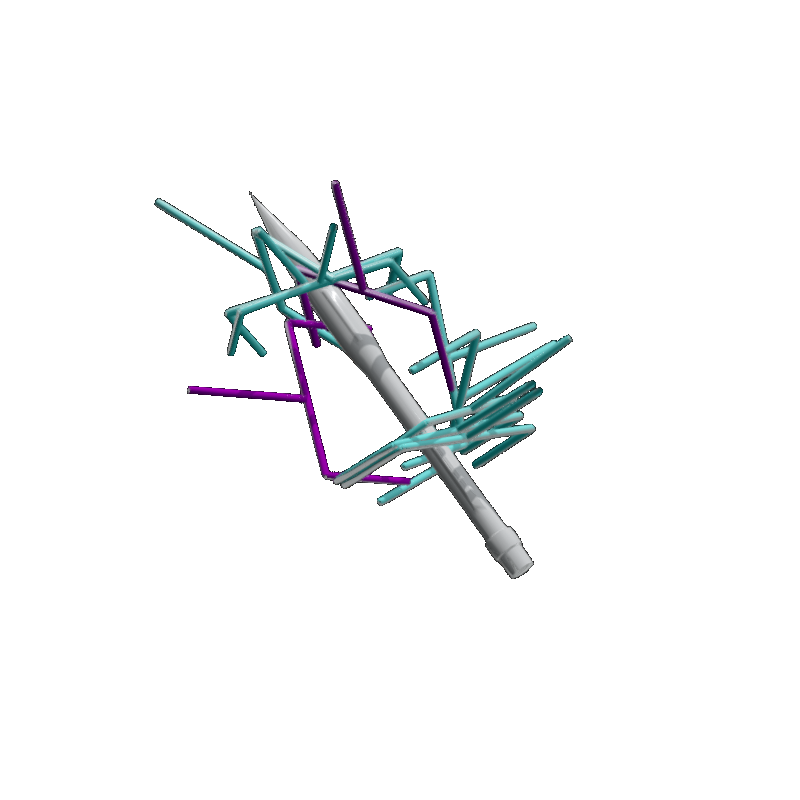}
        \end{overpic}
    \end{subfigure}
    \hfill
    \begin{subfigure}{\grenderwidthpencil}
        \begin{overpic}[width=\textwidth,tics=10,trim=120 120 120 50,clip]{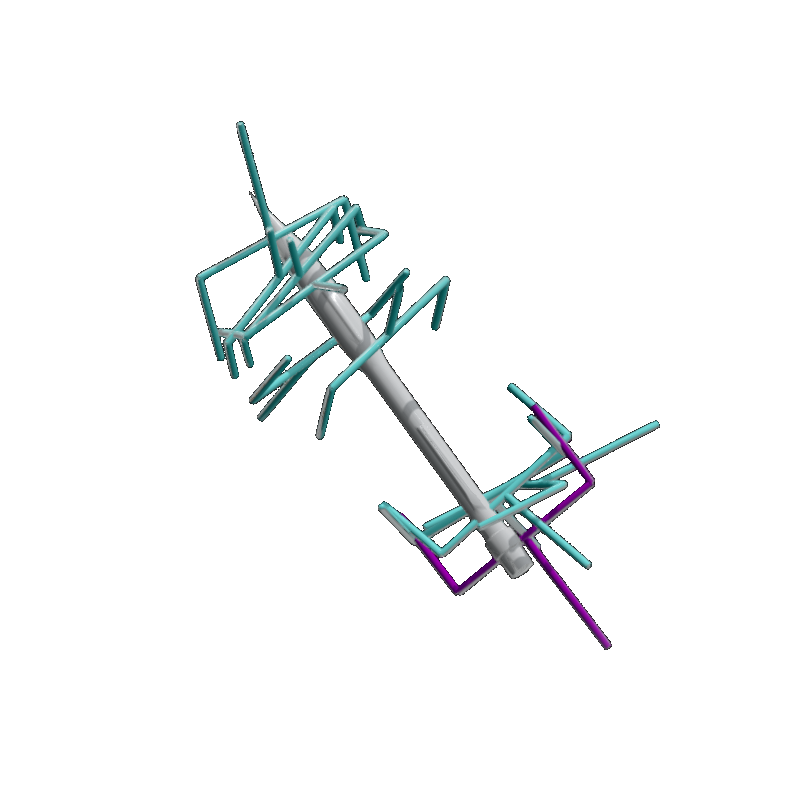}
        \end{overpic}
    \end{subfigure}
    \hfill
    \begin{subfigure}{\grenderwidthpencil}
        \begin{overpic}[width=\textwidth,tics=10,trim=120 120 120 50,clip]{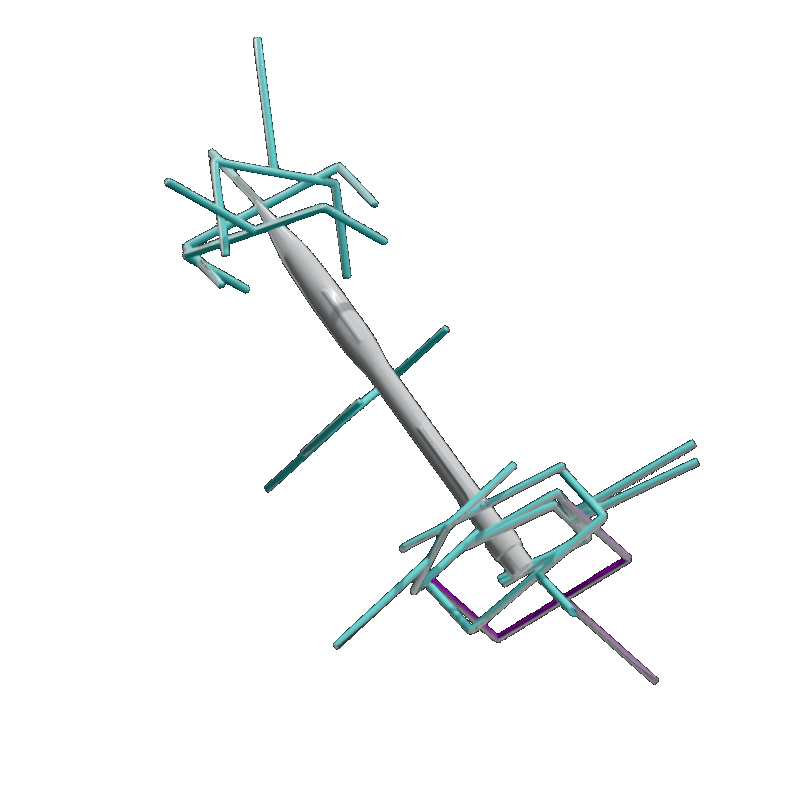}
        \end{overpic}
    \end{subfigure}
    \hfill
    \begin{subfigure}{\grenderwidthpencil}
        \begin{overpic}[width=\textwidth,tics=10,trim=120 120 120 50,clip]{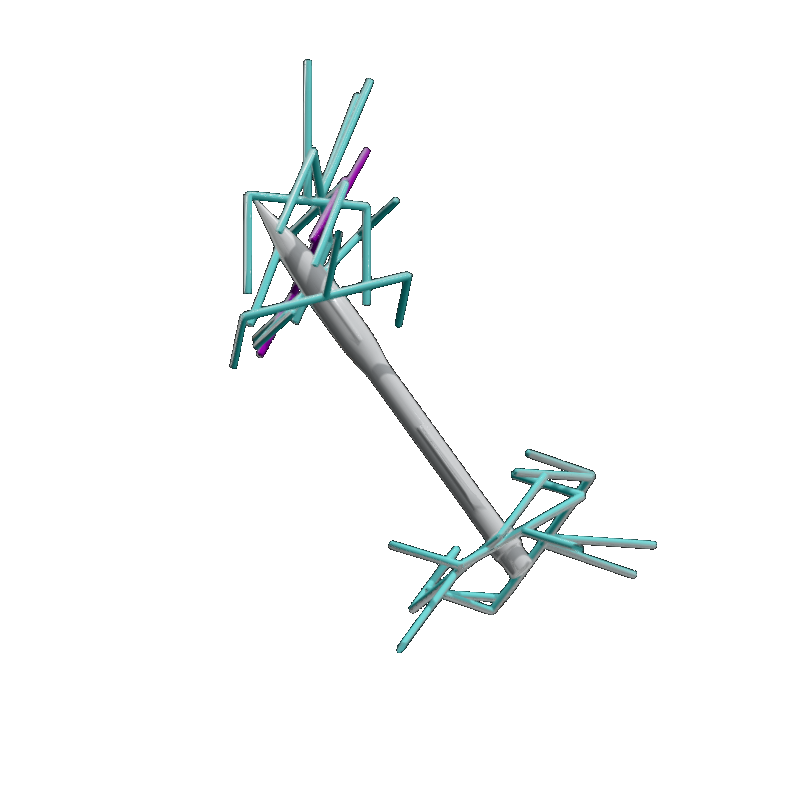}
        \end{overpic}
    \end{subfigure}
    \caption{\label{fig:grasp-renders}
    Qualitative comparison of generated grasps for Laptop (\emph{top row}), Donut (\emph{middle row}), and Pencil (\emph{bottom row}).
    \textcolor{grasp1}{\bf Dark cyan} indicates successful grasps while \textcolor{grasp0}{\bf dark purple} indicates failures.
    Compared to baselines, our method generates more successful and stable grasps across diverse object geometries. %
    }\vspace{-2mm}
\end{figure*}

\begin{figure}%
    \centering
    \begin{subfigure}{0.49\columnwidth}
        \includeinkscape[width=1.01\columnwidth]{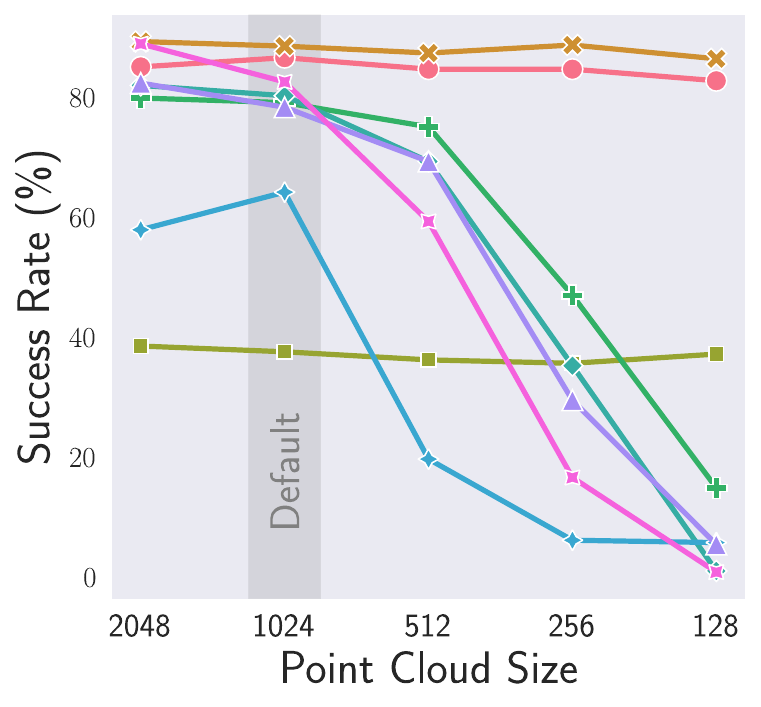}
        \label{fig:sclb-sr}
    \end{subfigure}
    \begin{subfigure}{0.49\columnwidth}
        \includeinkscape[width=1.01\columnwidth]{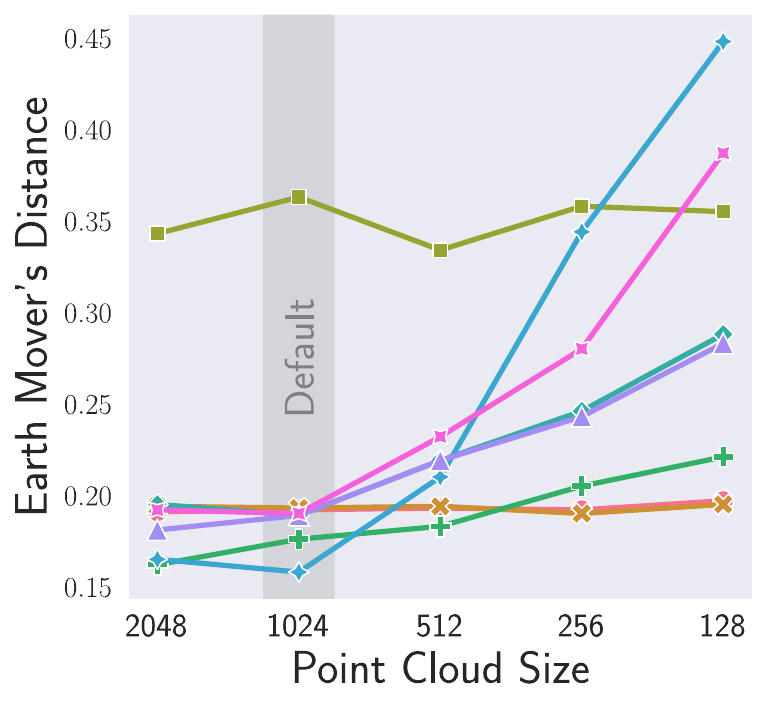}
        \label{fig:sclb-emd}
    \end{subfigure}
    \newline\vspace{-9mm}
    \begin{subfigure}{\columnwidth}
        \includeinkscape[width=\columnwidth]{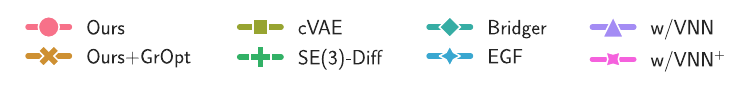}
        \label{fig:sclb-legend}
    \end{subfigure}
    \vspace{-10mm}
    \caption{\label{fig:sclb}
    Robustness evaluation under point cloud sparsity.
    Baselines using standard PointNet-based encoders show deteriorating success rate (SR) and diversity (EMD) with increasing sparsity.
    Our variational shape inference approach maintains consistent performance across all sparsity levels.
    }\vspace{-3mm}
\end{figure}

We evaluate on 52 unseen object instances from the ACRONYM dataset, measuring average success rate (SR) and an empirical Earth Mover's Distance (EMD) with respect to the ground truth distribution as indicators of grasp quality and diversity, respectively.
\cref{fig:boxplots} presents our main results.

The cVAE baseline suffers due to training on a limited set of objects for grasp synthesis, while also demonstrating a lack of diversity due to mode collapse~\cite{urain2023se}.
Furthermore, to boost its performance the cVAE depends on guidance from a grasp classifier~\cite{mousavian20196} which is expensive to train.
GLDM's emphasis on learning a latent grasp space, which we hypothesize is not needed due to already working in a low dimension, proves less effective.
Although the SMLD formulation of $\SE3$-Diff achieves reasonable performance, we believe it is limited by the VNN-based shape encoder.
Bridger improves on $\SE3$-Diff by employing stochastic interpolants, achieving a higher SR with significantly lower inference time.
EGF uses flow matching to achieve fast sampling, but generates less feasible grasps, which we attribute to requiring longer training times to handle the challenging set of objects.
Our method achieves the highest success rate of $86.5\%$, outperforming the previous state-of-the-art by $6.3\%$, while maintaining competitive diversity (EMD = $0.192$) and reasonable inference time ($\lesssim2$ seconds).

Key to our performance is an improved grasp synthesis architecture, a robust latent space for shape feature learning, and pre-training the shape backbone for object reconstruction.
To see that the improved grasp synthesis architecture alone does not contribute to this performance improvement, we train "w/VNN" which yields lower performance.
We also show the usefulness of pre-training for learning shape features by training "w/VNN$^+$", where the VNN is pretrained via a surface reconstruction objective (described in \cref{ssec:recon}).
Doing so boosts grasp synthesis performance compared to "w/VNN" but underperforms compared to Ours.
This observation is supported by the shape inference results in~\cref{ssec:recon}.
Finally, our proposed test-time pose optimization method (Ours + GrOpt) boosts success rate to $88.5\%$ by encoding collision avoidance and pinch stability objectives.
While this optimization slightly reduces diversity due to local refinement, it improves grasp quality and reduces variance.
Qualitative results in \cref{fig:grasp-renders} illustrate the higher grasp quality achieved by our method across a diverse set of objects.

\subsection{Sparse/Partial Point Clouds}
\label{ssec:pc-sparse}

Real-world perception systems often produce sparse or noisy point clouds due to sensor limitations and environmental factors.
We evaluate robustness by testing on point clouds of varying density, from 128 to 2048 points, where the training default is 1024.
We exclude GLDM in this analysis as it does not ingest point clouds of varying sizes.
\cref{fig:sclb} shows that methods using standard point cloud encoders experience significant performance degradation as point cloud sparsity increases.
In contrast, our approach has consistent performance across all sparsity levels, with success rates remaining above $80\%$ even with highly sparse inputs.
This robustness stems from our variational shape inference approach, which learns a structured latent space that captures geometric priors and compensates for sparsity.

The cVAE baseline~\cite{mousavian20196} also shows robustness due to its latent encoding, though at much lower performance.
Notably, the pre-trained VNN encoder (w/VNN$^+$) shows greater sensitivity to point cloud size than the non-pre-trained version, suggesting overfitting to the training point cloud density.
Our test-time optimization maintains consistently high performance across all sparsity levels, demonstrating its effectiveness even with limited shape measurements.

In many practical scenarios, using the full surface point cloud is not feasible.
We evaluate our method's efficacy in such scenarios by training on partial surface point clouds rendered from arbitrary camera view-points centered on the object.
During training, we filter ground truth grasps for which the swept volume of the gripper does not contain any point from the rendered point cloud, i.e., only grasps that encapsulate the partial point cloud.
Results are presented in \cref{fig:boxplots-partial}, where our method outperforms the baselines at $75.8\%$ success rate while abiding by the ground truth grasp distribution for the corresponding partial point cloud.

\begin{figure}[t]
    \centering%
    \includegraphics[width=0.7\columnwidth]{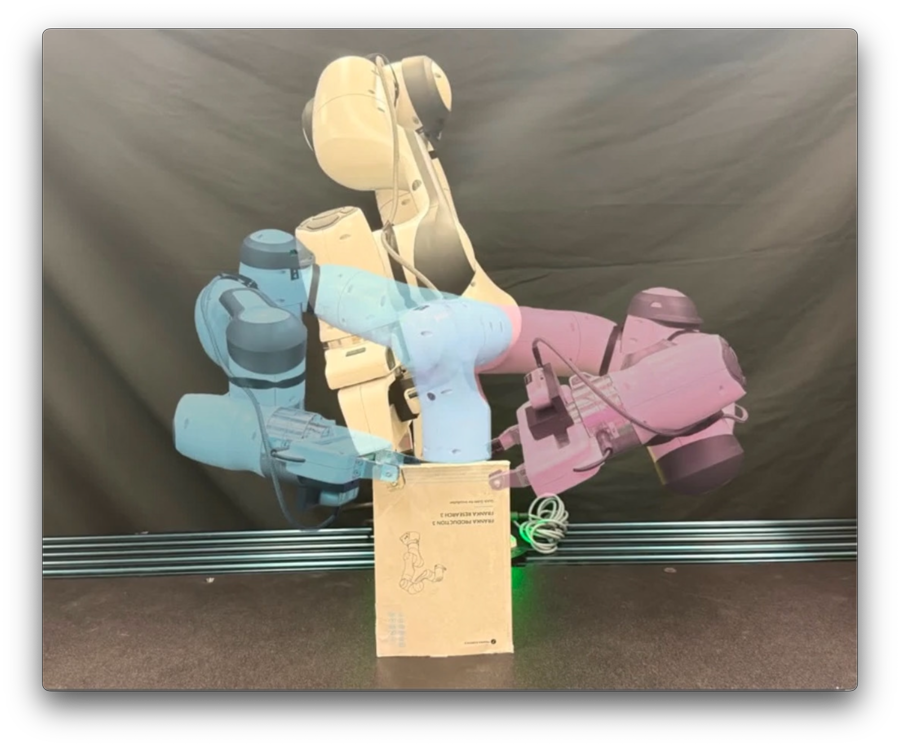}
    \vspace{-1mm}
    \caption{\label{fig:rw}
    Real-world grasp execution on a book using the Franka Research 3, demonstrating successful zero-shot transfer from simulation to real-world manipulation despite measurement noise and point cloud registration errors.
    }
    \vspace{-3mm}
\end{figure}

\subsection{Real-world Deployment}

To validate the effectiveness of our variational shape inference approach with real sensor data, we deploy our method on a Franka Research 3 robot arm equipped with a Panda Hand gripper in a pick-and-place task.
The experimental setup uses a Realsense D435i camera to capture the robot's workspace, with segmented object point clouds extracted using SAM~\cite{kirillov2023segany} as shown in \cref{fig:rw}.
Collision-free trajectories to and from the generated grasp poses are computed using the RRT-Connect implementation from VAMP~\cite{vamp_2024}.
Video demonstrations are provided in the supplementary material.

All methods are deployed without additional training or fine-tuning, evaluating zero-shot sim-to-real transfer capabilities.
As presented in \cref{table:rw}, our method achieves consistently high performance across all test objects, with particularly robust results for challenging geometries such as the Jug and Klein Bottle (K-Bottle).
The Jug is significantly larger compared to other objects, and the Klein Bottle features smoothly varying curvature and complex geometry.
Furthermore, real-world factors such as sensor noise and point cloud registration errors exacerbate the difficulty for methods that rely on standard point cloud encoders, while highlighting the advantage of our variational shape inference approach (trained only on noise-free point clouds).

\section{Conclusion}

In this work, we bring together approaches from generative modeling and implicit neural representations to enable robust and generalizable grasp synthesis.
We employ a shape inference formulation parameterizing SDFs conditioned on object point clouds, and use the learned model to formulate grasp synthesis via a denoising diffusion model.
Our proposal achieves superior performance compared to the state of the art in multimodal grasp synthesis, and demonstrates robustness in the presence of observation noise.
Additionally, our test-time optimization technique further improves grasp quality without additional training on expert data.

This work furthers the case for the usefulness of transferring geometry understanding approaches to grasp synthesis~\cite{jiang2021synergies}.
The demonstrated robustness to point cloud sparsity and the successful real-world experiments underscore the practical value of variational shape inference for robotic manipulation.
Therefore, in the future, large pretraining for shape inference may immediately be made useful for grasp synthesis in a few-shot manner~\cite{vosylius2024instant}.

A bottleneck in our presented approach is slower inference times.
This is mainly due to the repeated automatic differentiation calls during Langevin Dynamics.
However, we note that this does not pose a limitation to our shape inference formulation.
As such our approach may be coupled with faster diffusion-based sampling techniques \cite{chen2024behavioral}.

\begin{table}[t]
    \centering\vspace{1.5mm}%
    \small  %
        \begin{tabularx}{\columnwidth}{cccccc}\toprule
            \multirow[c]{2}{*}{Methods} & \multicolumn{5}{c}{Trials ($\cdot/10$)} \\ \cmidrule{2-6}
            & Book & Hammer & Jug & K-Bottle & Marker  \\
            \midrule
            cVAE~\cite{mousavian20196} & 0 & 0 & 0 & 0 & 0 \\
            SE(3)-Diff~\cite{urain2023se} & 10 & 5 & 0 & 0 & 9 \\
            Bridger~\cite{chen2024behavioral} & 9 & 4 & 3 & 1 & 2 \\
            GLDM~\cite{barad2024graspldm} & 10 & 4 & 0 & 1 & 0 \\
            EGF~\cite{lim2024equigraspflow} & 7 & 4 & 0 & 0 & 4 \\
            \midrule
            \hicell Ours & 10 & 8 & 9 & 6 & 8 \\
            \bottomrule
        \end{tabularx}
    \caption{\label{table:rw}
    Zero-shot sim-to-real grasp synthesis results on household objects.
    Our approach demonstrates consistent performance across objects with varying geometries despite sensor noise and calibration errors in the real-world setting.
    K-Bottle denotes the Klein Bottle.
    }
    \vspace{-4.5mm}
\end{table}

\renewcommand*{\bibfont}{\footnotesize}
\bibliographystyle{IEEEtranN}
\bibliography{IEEEabrv,references}

\end{document}